\documentclass[nohyperref,11pt]{article}

\usepackage[margin=1in]{geometry}
\usepackage{mathpazo}
\usepackage{amsmath,amssymb,amsthm,amsfonts,bm}
\usepackage{mathtools}
\usepackage{thmtools}
\usepackage{xspace}
\usepackage{graphicx}
\usepackage[dvipsnames]{xcolor}
\usepackage{booktabs}
\usepackage[colorlinks=true,linkcolor=blue,citecolor=blue,urlcolor=blue]{hyperref}

\usepackage[normalem]{ulem}
\usepackage{subcaption}
\usepackage{todonotes}
\usepackage[toc,page,header]{appendix}
\usepackage{minitoc}
\usepackage{url}
\usepackage{comment}
\usepackage{algorithm} %
\usepackage[toc,page,header]{appendix}
\usepackage{wrapfig}
\usepackage{multirow}
\usepackage{multicol}
\usepackage[T1]{fontenc}
\usepackage[backend=biber,style=alphabetic,natbib=true,maxbibnames=99,minalphanames=3]{biblatex}
\usepackage{cleveref}
\usepackage{enumitem}
\usepackage{soul}
\setuldepth{Hi}

\usepackage[noend]{algpseudocode}

\creflabelformat{equation}{#2\textup{#1}#3}

\definecolor{mydarkblue}{rgb}{0,0.08,0.85}
\definecolor{mylightblue}{rgb}{0.06,0.56,1.0}
\definecolor{mylightorange}{rgb}{1.0,0.62,0.12}
\definecolor{mylightred}{rgb}{0.99,0.00,0.04}
\hypersetup{colorlinks=true,
linkcolor=mydarkblue,
citecolor=mydarkblue,
filecolor=mydarkblue,
urlcolor=mydarkblue,
bookmarksopen=true,
linktoc=page}

\newtheorem{remark}{Remark}

\newtheorem{definition}{Definition}

\newcommand{\methodsc}{\textsc{Coar}\xspace}
\newcommand{\editsc}{\textsc{Coar-Edit}\xspace}

\newcommand{\skipthis}[1]{\ignorespaces}

\usepackage[normalem]{ulem}
\usepackage{xcolor}

\newcommand{\addlm}[1]{#1}

\newcommand{\scrcmd}{\textsuperscript}
\newcommand{\scr}[1]{\scrcmd{#1}}

\title{\methodsc}
\title{Decomposing and Editing Predictions \\ by Modeling Model Computation}

\author{
    Harshay Shah \\
    \texttt{harshay@mit.edu} \\
    MIT
    \and
    Andrew Ilyas \\
    \texttt{ailyas@mit.edu} \\
    MIT
    \and
    Aleksander M\k{a}dry \\
    \texttt{madry@mit.edu} \\
    MIT
}
\date{}

\begin{document}
\setcounter{tocdepth}{2}
\doparttoc %
\renewcommand\ptctitle{}
\faketableofcontents %

\maketitle
\begin{abstract}
\label{abs}
{\em How does the internal computation of a machine learning model transform
inputs into predictions?}
In this paper, we introduce a task called \emph{component
modeling} that aims to address this question. 
The goal of component modeling
is to decompose an ML model's prediction in terms of its {\em components}---simple
functions 
(e.g., convolution filters, attention heads)
that are the ``building blocks'' of model computation.
We focus on a special case of this task, {\em component attribution},
where the goal is to estimate the counterfactual impact of individual components 
on a given prediction.
We then present \methodsc, a scalable algorithm for estimating component attributions;
we demonstrate its effectiveness across models, datasets, and modalities.
Finally, we show that component attributions estimated with \methodsc directly enable model
editing {across five tasks, namely: fixing model errors, ``forgetting'' specific classes, 
boosting subpopulation robustness, localizing backdoor attacks, 
and improving robustness to typographic attacks}.
We provide code for \methodsc at \url{https://github.com/MadryLab/modelcomponents}.

\end{abstract}

\section{Introduction}
\label{sec:intro}
Despite their predictive power, large-scale
{machine learning (ML) models} remain black boxes.
In particular, the complex internal computation that these models
perform {to transform inputs into predictions} makes it difficult to
understand model behavior and,
as a result, detect failure modes prior to
deployment~\cite{beery2018recognition,sheng2019woman,geirhos2020shortcut}.

In response to this difficulty, a line of work in {ML} interpretability aims to shed
light on model computation by analyzing \emph{model components}---intuitively ``grouped''
model parameters such as convolutional filters or attention heads.
For example, feature visualization methods \cite{simonyan2013deep,zeiler2014visualizing,ghiasi2022vision} identify components in vision models that detect visual concepts such as curves \cite{olah2020overview} and objects \cite{bau2020understanding}.
Representation-based probes~\cite{alain2016understanding}
identify groups of components in
language models that encode sentiment~\cite{radford2017learning}, part-of-speech
tags~\cite{blevins2018deep}, and syntactic structure~\cite{hewitt2019designing}.
Finally, mechanistic
interpretability analyses~\cite{wang2022interpretability,nanda2023progress}
uncover specific components that encode a model
behavior of interest, e.g., ``knowledge neurons''~\cite{dai2021knowledge} and
``induction heads''~\cite{olsson2022in}. %
Broadly, these works leverage different tools to answer the question:
\emph{How do individual model components shape model behavior?}

In this work, we propose a new (and complementary) approach
to studying this question.
Our point of start is to rephrase the question, instead asking:
\begin{center}
  {\em
  How do changes to model components
  collectively change individual model predictions?
  }
\end{center}
We turn this rephrased question into a concrete task called
component modeling. In this task, the goal is to build
an interpretable {\em counterfactual estimator}
of how a model's output would change in response
to interventions made to its components.
In the rest of the paper,
we present a general approach to building such {estimators},
which turn out to be highly predictive in large-scale settings.
{Beyond shedding light on how model components collectively
contribute to a given prediction, these estimators enable effective
model editing, allowing us to design targeted interventions that
induce certain desirable model predictions.}

\begin{figure}
  \centering
  \includegraphics[width=\textwidth]{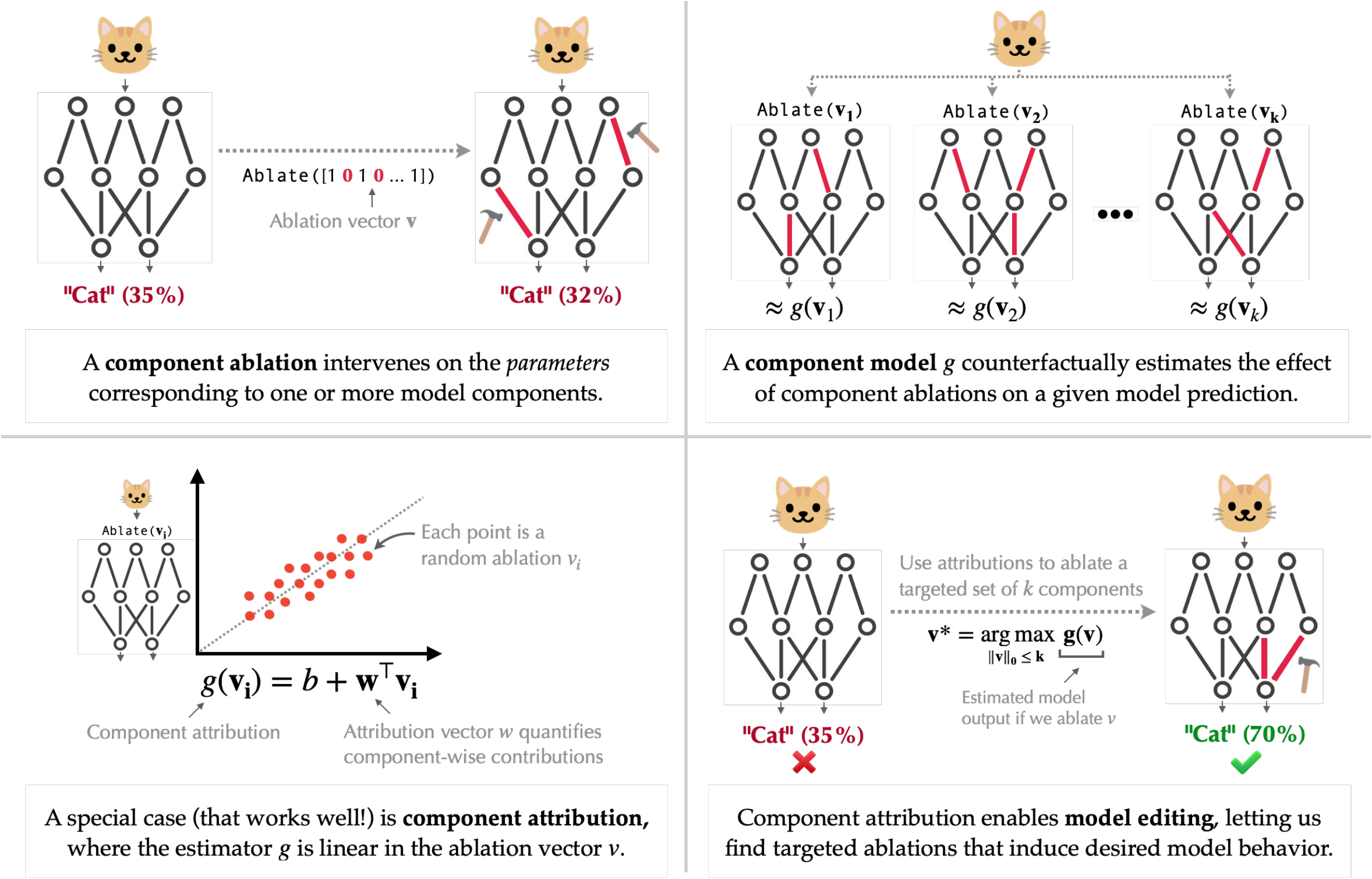}
  \caption{A summary of the component modeling framework.}
  \label{fig:headline}
\end{figure}

\paragraph{Roadmap \& contributions.}
The main contribution of our work is a framework for decomposing model
predictions in terms of model components, which we show has direct applications
to model editing.
Figure \ref{fig:headline} summarizes these contributions.
Specifically, in this paper we:
\begin{enumerate}
    \item
    {\bf Introduce the component modeling framework:}
    We formalize our goal of understanding how model components
    shape ML predictions as a concrete task called
    {\em component modeling} (\Cref{def:problem}).
    {The objective of this task is to learn a counterfactual estimator, or
    \emph{component model}, that accurately predicts the effect of ablating a
    subset of model components on a given model prediction (\Cref{eq:ablation}).}
    Intuitively, this task operationalizes the idea that if we can ``understand'' how model components shape a prediction, we should also be able to
    {\skipthis{counterfactually} estimate} how the prediction would change if we were
    to ablate a subset of components.
    \item {\bf Instantiate the framework via component attribution:}
    We focus our attention on a special ``linear'' case of component modeling called
    {\em component attribution},
    where we assign a score to each \skipthis{model} component, and estimate the counterfactual effect of
    ablating a set of components as the sum of their corresponding scores (\Cref{def:attribution}).
    Component attributions allow us to directly read off the ``contribution'' of
    every component to a prediction, abstracting away the complexity of
    the model's internal computation.
    \item {\bf Propose an algorithm for efficient component attribution:}
    We develop \methodsc (\underline{co}mponent \underline{a}ttribution via \underline{r}egression),
    a scalable way to estimate component attributions (\Cref{sec:method}).
    Through experiments on {both image classifiers and language models}, we show that \methodsc yields component attributions that can accurately predict how model predictions change in response to component-level ablations (\Cref{sec:eval}).
    \item {\bf Demonstrate that component attributions enable model editing:}
    Component attributions from \methodsc
    directly {{enable} edits to large-scale classifiers without additional training} (\Cref{sec:applications}).
    Specifically, we propose an editing procedure (\editsc)
    that designs targeted ablations by using
    \methodsc attributions as a counterfactual estimator---given an objective,
    \editsc finds ablations for which estimated model outputs perform well.
    We stress-test \editsc through five editing tasks:
    fixing model errors (\S\,\ref{subsec:examples}),
    selectively ``forgetting'' an entire class (\S\,\ref{subsec:class_forget}),
    boosting subpopulation robustness (\S\,\ref{subsec:subpops}),
    localizing backdoor attacks (\S\,\ref{subsec:backdoor}), and
    mitigating typographic attacks (\S\,\ref{subsec:typographic}).
\end{enumerate}
\paragraph*{Paper organization.}
We begin by formalizing the component modeling task and its special case, component attribution, in \Cref{sec:problem}.
We then describe our method, \methodsc, for estimating component attributions in \Cref{sec:method}, and demonstrate its effectiveness on large-scale models in \Cref{sec:eval}.
Finally, we stress-test the practical utility of \methodsc attributions for model editing in \Cref{sec:applications}.

\section{Setup and Problem Statement}
\label{sec:problem}
Consider a typical supervised learning setup.
We have a set $S$ of input-label pairs (or {\em examples}) $z_i=(x_i, y_i)$,
and a trained model $M$ that maps inputs $x$ to predicted labels
$M(x)$.
We define the \emph{model output function}
$f_M(z) \in \mathbb{R}$ as any statistic that quantifies the correctness of
model $M$ on the example $z$. %
For instance, the model output $f_M(z)$ can be the cross-entropy loss in a classification task, or the squared loss in a regression task. %

In this work, we will think of the model
$M$
not as a black box,
but instead as the output of a {\em computation graph} $G_M$ \citep{bauer1974computational}.
Each parameterized node of this graph---which we call a {\em component}---is a
function mapping its incoming edges to an outgoing edge.
For example, a $d$-dimensional linear model $M$ naturally admits
a computation graph $G_M$ with $d$ components---one component
$C_i(z) = w_i \cdot z$ for each parameter $w_i$---followed
by a summation that combines the components into an output.

For more complex models,
there are often multiple valid computation graphs $G_M$ we could consider.
For example, if $M$ is a Transformer \citep{vaswani2017attention},
the components might be multi-head attention blocks,
individual attention heads, or even individual parameters.
In general, the component set depends on
the model architecture {\em and} the level of granularity that we wish to study.

\paragraph{Component modeling.}
Our goal in this work is to understand the behavior of the model $M$ in terms of its {components}.
By viewing the model $M$ as a computation graph $G_M$ over a set of components
$C$, we can restate our goal as follows:
\begin{center}
    \emph{Given a model $M$ and example $z$, how do individual components $c \in C$ \\ combine to yield the model output~$f_M(z)$?}
\end{center}
Of course, there is a trivial answer to this question:
the components $c \in C$ combine
through the very computation graph used to define $C$.
This answer is correct but not satisfying, as it does not get us closer to our conceptual goal of
understanding model behavior in terms of components.

What we are truly after is
a {\em simple, interpretable} function capturing how components in $C$ impact $f_M(z)$.
To make this more precise, we define the
{\em component counterfactual function}
$f_M(z, C')$ as
\begin{equation}
    \label{eq:ablation}
    f_M(z, C') := \text{model output } f_M(z) \text{ on example } z \text{ \emph{after} ablating components } C' \subseteq C,
\end{equation}
where ``ablating'' here corresponds to any {intervention} that
overrides or patches the {parameters corresponding to} components $c \in C'$ (e.g., by setting them
to zero~\cite{olsson2022in} or by adding random noise~\cite{meng2022locating}).

Equation \eqref{eq:ablation} allows us to
operationalize our goal as a
\emph{counterfactual estimation} task.
In this task, we want to estimate component counterfactuals $f_M(z,C')$
using a much simpler function, which we call a component model.

\begin{definition}[Component modeling]
    \label{def:problem}
    Fix a model $M$
    with computation graph $G_M$, component set $C = \{c_1,\ldots,c_N\}$,
    and model output function $f_M$.
    For any subset of model components $C' \subseteq C$,
    let $\bm{0}_{C'}$ be the corresponding ablation vector of $C'$,
    defined as a $N$-dimensional vector where
    \begin{align*}
        (\bm{0}_{C'})_i = \begin{cases}
        0 & \text{if } c_i \in C' \\
        1 & \text{otherwise.} %
        \end{cases}
    \end{align*}
    Given an example $z$, a \underline{\smash{component model}} for $z$
    is a function
    $\smash{g^{(z)}}: \{0, 1\}^N \to \mathbb{R}$
    that maps ablation vectors of subsets $C'$ to
    estimates of the counterfactual $f_M(z, C')$.
\end{definition}
\noindent
In other words, the high-level goal of component modeling is to build an {estimator}
that can directly {answer} {counterfactual questions} like
{\em ``what would happen to my classifier's prediction on a given image if I ablated a
specific set of components $C' \subseteq C$?''}
without having to intervene on the computation graph $G_M$ and ablate components in $C'$.

\paragraph{Component attribution.}
In this work, we consider a subcase of component modeling---which
we call {\em component attribution}---where the function $g^{(z)}$ is {\em
linear} in its input.
That is, a component attribution
{for example $z$}
assigns a {score} $\smash{\bm{w}_i^{(z)}}$ to each component $c_i \in C$,
and predicts the effect of ablating $C' \subset C$ as
the sum of the scores corresponding to components in $C \setminus C'$. %
\begin{definition}[Component attribution]
    \label{def:attribution}
   Fix a model $M$ with output function $f_M$ and
   component set $C = \{c_1, \ldots, c_N\}$.
   A \ul{component attribution} for example $z$ is a
   set of parameters $\smash{\bm{\theta}^{(z)} \coloneqq
   \{\bm{w}^{(z)} _1\!\!,\ldots,
   \bm{w}^{(z)}_N,b^{(z)}\}}$
   which parameterize a linear component model, i.e., a function $g^{(z)}$
   such that
   \[
    f_M(z;C') \approx g^{(z)}(\bm{0}_{C'}) \coloneqq \bm{0}_{C'}^\top \bm{w}^{(z)} + b^{(z)}
   \]
\end{definition}
\noindent Component attribution satisfies our goal of finding a simple,
interpretable account of how model components combine to form
predictions.
In particular, a component attribution for example $z$
decomposes a model's output on
$z$ into the contributions $\smash{\bm{w}_i^{(z)}}$
of each individual component $c_i$.

\begin{remark}[Linearity and misspecification]
\label{rem:misspec}
Modern ML models comprise complex computation graphs with highly non-linear
interactions among model components.
For such models, it is unclear a priori why the effect of ablating
components on model outputs (i.e., component counterfactuals \eqref{eq:ablation})
should be well-approximated by {\ul{linear}} component attributions
\eqref{def:attribution}, which sum
fixed additive effects of individual components.
Still, despite this evident misspecification,
our results on large-scale vision and language models in \Cref{sec:eval}
show that component attributions {\ul{can}} accurately predict component
counterfactuals.
\end{remark}

\section{Component attribution with \methodsc}
\label{sec:method}
In~\Cref{sec:problem}, we formalized our high-level goal of
understanding how models internally process examples into a
counterfactual estimation task called {\em component modeling}
(\Cref{def:problem}),
of which we study a special (linear) case called {\em component attribution}
(\Cref{def:attribution}).
Now, we show how to estimate component attributions
$\bm{\theta}^{(z)}$ by casting the counterfactual estimation task as a regression problem.
Specifically, we now describe \methodsc (\underline{co}mponent \underline{a}ttribution via \underline{r}egression),
a {general} component attribution method for models ranging from
random forests to {deep} neural networks.

\paragraph{Approach.}
Consider a \skipthis{given (i.e., fixed)} fixed model output $f_M(\cdot)$ of interest,
and a corresponding computation graph $G_M$ that encodes
the model components $C$ at the desired level of granularity.
Additionally,
we fix an {\em ablation method}, i.e.,
a procedure for ``overriding'' or patching any given subset $C' \subset C$ of the model
components in the computation graph $G_M$.

Our method \methodsc takes in an example $z$
and outputs a corresponding component attribution vector
$\bm{\theta}^{(z)} \in \mathbb{R}^{|C|+1}$
(\Cref{def:attribution}).
{To do so}, \methodsc casts the task of predicting component counterfactuals
as a \emph{supervised learning} problem,
which we \skipthis{then} solve in {two steps}:

\begin{enumerate}
    \item \textbf{Construct a component dataset.}
We construct a dataset $D^{(z)}$ of
component counterfactuals for the example $z$.
Each ``datapoint'' in $D^{(z)}$ consists of
a component subset $C_i \subseteq C$ and
its corresponding counterfactual $f_M(z, C_i)$
(see \eqref{eq:ablation})---we evaluate the latter
by simply ablating the components in $C_i$ and evaluating the model on
example $z$.

In this work, we choose the component subsets
$C_i$ to be random $\smash{\alpha_{\text{train}}}$-fraction subsets of the component set $C$,
for a ablation fraction hyperparameter $\smash{\alpha_{\text{train}} > 0}$.\footnote{
We opt for this random $\alpha$-fraction sampling method for simplicity---it
may be possible to make \methodsc more statistically
efficient by choosing the subsets $C_i$ more carefully.}
The output of this step is a {\emph{component dataset}}
\begin{align}
    \label{eq:dataset}
    D^{(z)} &:= \{(C_1, f_M(z, C_1)), \dots, (C_m, f_M(z, C_m))\},
\end{align}
where $C_i \sim \text{Uniform}(\{C' \subset C \text{ : } |C'| = \alpha_{\text{train}} |C| \})$.
We study the effect of varying the ablation fraction $\alpha$ \skipthis{(i.e., the hyperparameter $\alpha$)}
on \methodsc in \Cref{app:alpha}.
\item \textbf{{Fit} {a linear estimator}.} We then
use the dataset $D^{(z)}$ to
fit component attribution parameters
$\bm{\theta}^{(z)}$ for each example $z$
(see \Cref{def:attribution}).
More specifically, for each example $z$,
we minimize the squared loss between the component counterfactuals
from Step 1 and their corresponding attribution-based predictions by
solving the following {\emph{linear regression}} problem:
\begin{equation}
    \label{eq:regression}
    \bm{\theta}^{(z)} \coloneqq \arg\min_{b \in \mathbb{R},\,\bm{w} \in \mathbb{R}^{|C|}}
    \sum_{(C_i, f_M(z, C_i)) \in D^{(z)}} \bigg(
        b + \bm{0}_{C_i}^\top \bm{w} - f_M(z, C_i)
    \bigg)^2,
\end{equation}
\noindent where again $\bm{0}_{C_i}$ is the ablation vector of $C_i$ (\Cref{def:problem}).
Our component model is then
\begin{equation}
    \label{eq:attribution_derived}
    g^{(z)}(\bm{0}_{C'}) \coloneqq \bm{0}_{C'}^\top \bm{w}^{(z)} + b^{(z)}.
\end{equation}
\end{enumerate}
We provide pseudocode for \methodsc in~\Cref{app:pseudocode}.
As we discussed in \Cref{sec:problem}, the resulting component attribution
$\bm{\theta}^{(z)}\coloneqq (\bm{w}^{(z)}, b^{(z)})$ is
interpretable in that the coefficient $\smash{\bm{w}_j^{(z)}}$ estimates how the model output on example $z$ would change if we were to ablate component $c_j$.
We can thus view this coefficient as the (estimated) additive contribution of component $c_j$ to the model output.

The above two-step approach is simple and highly scalable---we can
construct the dataset $D^{(z)}$ with
just forward passes on the given model to {compute component counterfactuals},
and optimize the linear regression problem (\cref{eq:regression}) with
off-the-shelf GPU-based solvers---see \Cref{app:implementation_details} for details.
This enables us to apply \methodsc on large-scale models (e.g., ViT \citep{dosovitskiy2021image})
and datasets (e.g., ImageNet \citep{deng2009imagenet}),
as shown in the {next} section.

\paragraph{Instantiating \methodsc for classifiers.}
{Our method \methodsc is \emph{general}} in that we can use it to study any
machine learning model $M$ that has a corresponding output function
$f_M$ and computation graph $G_M$.
{In this work, we primarily use \methodsc to analyze models trained on
classification tasks.}
Although the computation graph $G_M$ will vary based on the
specific model architecture we are studying, across all models we use
the standard \emph{correct-class margin}~\cite{ilyas2022datamodels}
as the model output $f_M$, i.e.,
\begin{equation}
    \label{eq:margin}
    f_M(z) := \text{(logit for correct class)} - \text{(highest logit for incorrect class).}
\end{equation}
a quantity whose sign indicates the correctness of model $M$ {on the example $z$}.
For the latter,
we choose to ablate component subsets $C' \subset S$
by simply setting the parameters of the components in $C'$
to zero~\cite{wang2022interpretability,olsson2022in}.
We use \methodsc with alternative model output functions and ablation methods in~\Cref{app:model_output_fn} and~\Cref{app:ablation_and_model_output} respectively.

\begin{remark}[Ablation is not removal]
\label{rem:ablation}
As noted in prior work \citep{chan2022causal},
ablation methods (e.g., setting weights or activations to zero)
do not ``remove'' model components from the computation
graph.
Instead, such ablations shift the activations off-distribution
in a {systematic} way---the goal of component attribution
(\Cref{def:attribution}) is to predict the change in model outputs induced by
this shift.
We use zero-ablation as
our ablation method because it is a  common choice in the
literature~\cite{olsson2022in,wang2022interpretability}.
In \Cref{app:ablation_and_model_output}, we show that \methodsc can estimate
component attributions with alternative ablation methods as well.
\end{remark}

\section{Does \methodsc learn accurate component attributions?}
\label{sec:eval}
We now evaluate whether \methodsc-estimated attributions
accurately predict component counterfactuals (as in \eqref{eq:ablation})
for deep neural networks trained on image classification and language modeling.

\paragraph{Datasets, models, and components.}
We apply \methodsc to compute component attributions in three
different setups:
\begin{itemize}%
    \item {\bf Setup A:} A ResNet-18~\cite{he2015deep} trained on the CIFAR-10
    dataset~\cite{krizhevsky2009learning}, with a computation graph $G_A$
    comprising $|C|=2,306$ components. Specifically, each model
    component $c_i \in C$ corresponds to a convolutional filter in the model,
    and ablating a set of components $C' \subset C$ means setting all the weights in
    the corresponding filters to zero.
    \item {\bf Setup B:} A ResNet-50 trained on the ImageNet
    dataset~\cite{deng2009imagenet}, with a computation graph $G_B$ comprising
    $|C| = 22,720$ components. Again, each component here corresponds to
    a convolutional filter in one of the $49$ convolution layers of the ResNet-50.
    \item {\bf Setup C:} A Vision Transformer
    (ViT-B/16)~\cite{dosovitskiy2021image} trained on ImageNet,
    whose computation graph $G_C$ comprises $82,944$ components.
    Each component here corresponds to a row of a weight matrix in one of $12$
    transformer blocks of the ViT, and ablating a set of components
    means setting the corresponding rows to zero.
\end{itemize}
We provide additional details on the models and datasets in \Cref{app:datasets_and_models}.

\paragraph{Applying \methodsc.}
We use \methodsc to {obtain} component attributions
(one for each test example)
in each setup.
Specifically, for a given model,
we first construct a component dataset $D^{(z)}$ for
each example $z$ (as in Step 1 of \Cref{sec:method}) by
randomly ablating $\alpha_{\text{train}}$ fraction of all components and evaluating the resulting correct-class margin \eqref{eq:margin} on $z$, where $\alpha_{\text{train}} = \{10\%,5\%,5\%\}$ for setup $\{\text{A}, \text{B}, \text{C}\}$ above.
We repeat this $m$ times, yielding a component dataset
$D^{(z)}$ of size $m$ for each example $z$---we
use
$m={\{50000,100000,200000\}}$ for setup $\{\text{A}, \text{B}, \text{C}\}$ above.
We then run linear regressions
on the resulting datasets (as in Step 2 of \Cref{sec:method}) to yield the final component attributions.
We defer implementation details to~\Cref{app:implementation_details} and study the effect of the dataset size $m$ and ablation fraction $\alpha_{\text{train}}$ on the resulting attributions in~\Cref{app:sample_complexity,app:alpha_eval}.

\paragraph{Evaluation metric.}
We evaluate component attributions based on
their ability to {estimate} {\em unseen} component counterfactuals \eqref{eq:ablation},
i.e.,
the result of ablating component subsets $C'$ not observed at training time.
Specifically,
we sample a new collection of $k$
component subsets
\[D_\text{test}^{(z)} \coloneqq \{C_1',C_2',\ldots,C_k\},
\qquad \text{ where }
\qquad
C_i' \sim \text{Unif}(\{C' \subset C: |C'| = \alpha_\text{test} |C|\}),
\]
where $0 < \alpha_\text{test} < 1$ is the ablation fraction used at evaluation time.
Varying $\alpha_\text{test}$ allows us to evaluate attributions on in-distribution ($\alpha_\text{test} = \alpha_\text{train}$) and out-of-distribution ($\alpha_\text{test} \neq \alpha_\text{train}$) component counterfactuals.

To quantify the predictiveness of component attributions, we use $\smash{D_\text{test}^{(z)}}$ to measure the Pearson correlation between component counterfactuals $f_M(z, C_i')$
and their corresponding attribution-based estimates $\smash{g^{(z)}(\bm{0}_{C_i'})}$ (\cref{eq:attribution_derived}.), i.e.,
\begin{equation}
    \label{eq:correlation}
    \rho(z) \coloneqq \text{Pearson-}\rho\Big(\underbrace{\{f_M(z, C_1'), ..., f_M(z, C_k') \}}_{\text{ground-truth \skipthis{component} counterfactuals}}, \underbrace{\{g^{(z)}(\bm{0}_{C_1'}), ..., g^{(z)}(\bm{0}_{C_k'}) \}}_{\text{attribution-based estimates}} \Big).
\end{equation}

\paragraph{Baselines.}
We use the evaluation metric described above (\Cref{eq:correlation}) to compare \methodsc attributions with four baselines, two adapted from related work and two natural baselines.
We defer implementation details to \Cref{app:baselines}.
\begin{itemize}%
    \item {\bf Adapted baselines (\texttt{NC}, \texttt{II})}:
    We adapt {\bf neuron conductance (\texttt{NC})} \cite{dhamdhere2018important}
    and {\bf internal influence (\texttt{II})} \cite{leino2018influence} to the
    component attribution setting. Both
    methods use integrated gradients~\cite{sundararajan2017axiomatic}
    (an input-space feature attribution method)
    to compute importance scores for each component $c_i$.
    To compare these methods to \methodsc, we apply \texttt{NC} and \texttt{II}
    to model outputs on example $z$,
    and interpret the resulting scores as the attribution coefficients $\smash{w_i^{(z)}}$.
    \item {\bf Specialized baselines (\texttt{LOO}, \texttt{GP}):} We also consider two other baselines.
    First, {\bf leave-one-out (\texttt{LOO})}
    ablates individual components $c_i$
    and estimates the corresponding coefficient based on the effect of the
    ablation, setting
    $\smash{\bm{w}^{(z)}_i =  f_M(z, \{c_i\}) - f_M(z, \emptyset)}$.
    We also consider {\bf gradient-times-parameter (\texttt{GP})}, {which approximates the leave-one-out effect
    of each component using a first-order Taylor approximation, setting
    $\smash{\bm{w}^{(z)}_i =
    \nabla_{c_i} f_M(z, \emptyset) \cdot \delta_{c_i}}$}, where $\delta_{c_i}$
    is the parameter-space change in $c_i$ induced by the ablation method of
    choice.
\end{itemize}

\begin{remark}[Relation between baselines and patching]
\label{rem:patching}
Readers familiar with the field of mechanistic interpretability may observe that the
leave-one-out (\texttt{LOO}) baseline resembles activation
patching~\cite{vig2020investigating,meng2022locating,zhang2023towards}
and the gradient-times-parameter (\texttt{GP}) baseline resembles
attribution patching~\cite{syed2023attribution,kramar2024atp*}.
Indeed, activation patching ablates individual activations to estimate their effect,
and attribution patching considers a first-order gradient-based approximation.
The key difference is that the ablations we consider are in \ul{parameter}
space rather than activation space, and that we study a different model output.
\end{remark}

\subsection{Results}
\begin{figure}[!t]
    \centering
    \includegraphics[width=0.99\textwidth]{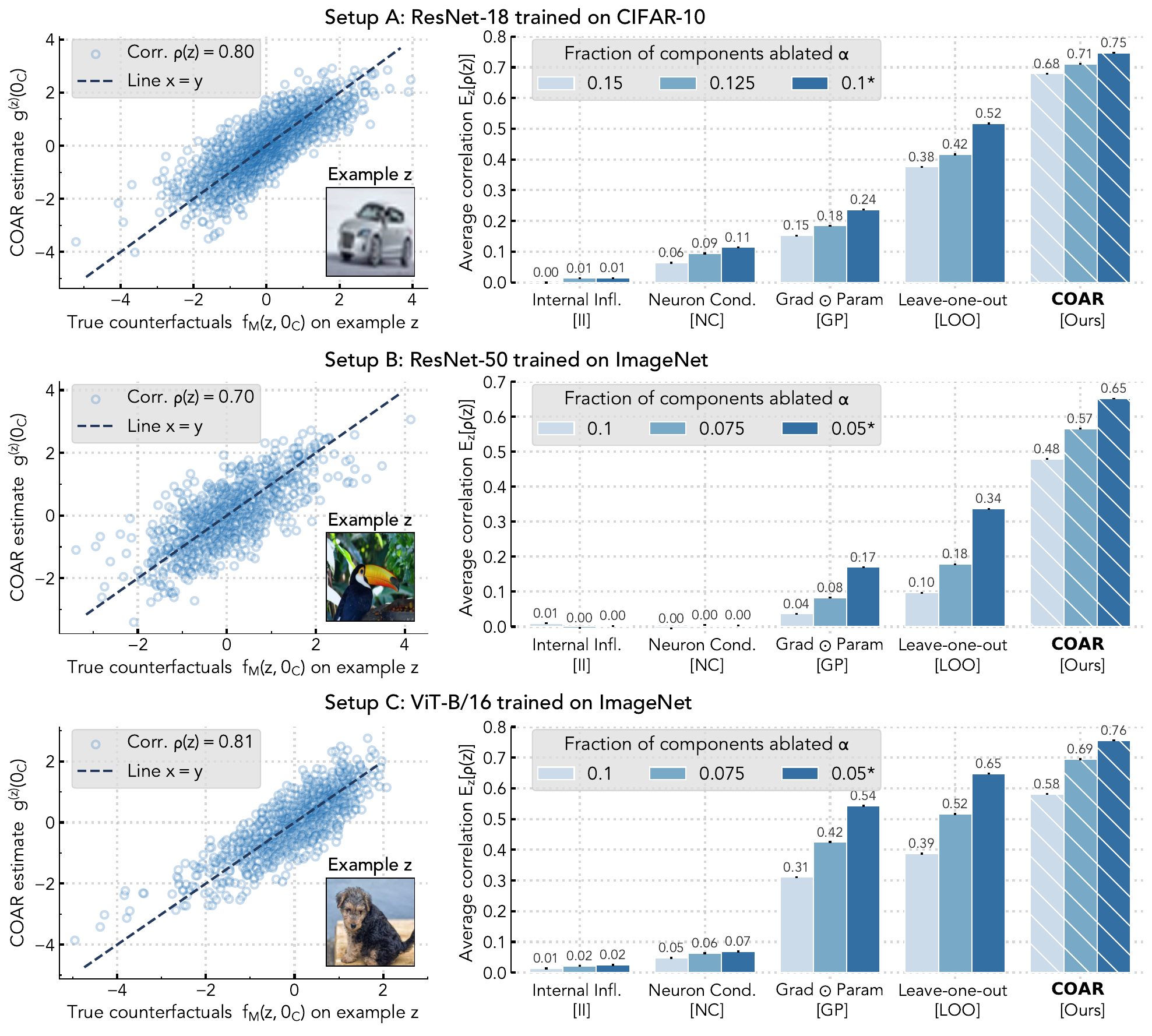}
    \caption{
        \textbf{Evaluating \methodsc attributions.}
        We evaluate whether component attributions computed using our procedure \methodsc
        accurately predict component counterfactuals~\eqref{eq:ablation}.
        We compare \methodsc to four baselines (described in \Cref{sec:eval}) on three image classification setups (one per row).
        The subfigures on the left each focus on a single example $z$ (visualized in the bottom-right corner of each plot),
        and show that for each setup, the ground-truth component counterfactuals
        $f_M(z, \cdot)$ ($x$-axis) and attribution-based estimates
        $g^{(z)}(\cdot)$ ($y$-axis) exhibit high correlation $\rho(z)$.
        On the right, we observe that \methodsc attributions exhibit high average correlation $\mathbb{E}_z[\rho(z)]$ over test examples,
        outperforming all baselines in each task and for all ablation fractions $\alpha_{\text{test}}$.
        The asterisk ($\mbox{*}$) in each legend denotes $\alpha_{\text{train}}$, the ablation fraction used to fit the
        component attributions.
    }
    \label{fig:eval}
\end{figure}

We now use the setup described above to test whether \methodsc learns accurate component attributions for setups $\{\text{A},\text{B},\text{C}\}$.
For each task, we first use \methodsc to estimate a component attribution for every example $z$ in the {corresponding} test set.
We then evaluate these component attributions using the correlation metric {$\rho(z)$} defined in \Cref{eq:correlation}.
\Cref{fig:eval} depicts our results. %

\paragraph{Example-level analysis.}
On the left side of each {row} in~\Cref{fig:eval},
we focus on an {\em individual test example}~$z$ from each task.
For each example $z$, we ablate random component subsets $C' \subset C$ of size
$\alpha_{\text{test}} \cdot |C|$ (for $\alpha_\text{test} = \alpha_\text{train}$)
from the model \skipthis{computation graph} and estimate the correlation $\rho(z)$
from \Cref{eq:correlation}.
Across all three tasks, we observe that \methodsc learns accurate component attributions for the selected test examples.
In \Cref{app:baseline_eval}, we provide additional (randomly selected)
example-specific correlation plots, as well as analogous plots for all
baselines described above.

\paragraph{Aggregate analysis.}
The right side of each {row} in~\Cref{fig:eval} plots the \emph{average} correlation between the ground-truth counterfactuals and attribution-based estimates over test examples, i.e., $\mathbb{E}_z[\rho(z)]$.
We also analyze the effect of ablation fraction $\alpha_\text{test}$ on the
average correlation, finding that:
\begin{enumerate}
    \item[(a)] \methodsc outperforms baselines by a large margin across datasets, models, and ablation fractions $\alpha_\text{test}$. For
    example, when \skipthis{randomly} ablating $\alpha_\text{test} = 5\%$ of components in the ImageNet ResNet50 \skipthis{model} (setup B),
    attribution-based estimates using \methodsc and the best-performing baseline (\texttt{LOO}) exhibit $0.65$ and $0.34$ correlation with \skipthis{the} ground-truth \skipthis{component} counterfactuals, respectively.
    Additionally, the adapted baselines, \texttt{NC} and \texttt{II}, exhibit low correlation with the ground-truth counterfactuals in all three setups.
    \item[(b)] The correlation between ground-truth counterfactuals and attribution-based estimates decays gracefully on larger out-of-distribution component subsets, i.e., as $\alpha_{\text{test}}$ increases. For example, increasing $\alpha_\text{test}$ from $10\%$ (equal to $\alpha_\text{train}$)
    to $12.5\%$ and $15\%$ on CIFAR-10 (setup A) only decreases the average correlation of \methodsc-based estimates from $0.74$ to $0.70$ and $0.68$ respectively. %
\end{enumerate}

\paragraph{Applying \methodsc to language models.}
Although we focus on vision models in this work, our attribution method \methodsc is general and modality-agnostic.
In~\Cref{app:language_models}, we show that \methodsc, without any
modification, yields predictive component attributions for language models as well.
First, we apply \methodsc to GPT-2~\cite{radford2019language} evaluated on the next-token prediction task using the TinyStories dataset~\cite{eldan2023tinystories} (\S\ref{app:gpt2}).
Then, we turn to the zero-shot classification setting and apply \methodsc to Phi-2~\cite{li2023textbooks} evaluated on the BoolQ question-answering task~\cite{clark2019boolq} (\S\ref{app:phi2}).

\paragraph{Additional analysis.} In~\Cref{app:additional_evaluation}, we show
that \methodsc attributions are predictive for
out-of-distribution inputs (\S\ref{app:dataset_eval}),
additional architectures (\S\ref{app:arch_eval}),
additional tasks (\S\ref{app:setting_eval}),
and different train-time ablation fractions (\S\ref{app:alpha_eval}).
We also show that \methodsc outperform \skipthis{exhibit higher correlation with ground-truth counterfactuals than} baselines when trained with $2$-$5\times$ fewer samples in~\Cref{app:sample_complexity}, and provide qualitative analysis in~\Cref{app:qualitative}.

\section{Do \methodsc attributions enable model editing?}
\label{sec:applications}
In the last section, we showed that \methodsc attributions accurately predict
how model outputs change in response to component-level
interventions. %
We now evaluate the practical utility of
\methodsc by applying it to the problem of \emph{model editing}.
{That is, we ask:}
\begin{center}
    \textit{Is ablating model components identified via \methodsc attributions an effective way to edit models?}
\end{center}
{To answer this question},
we first define model editing in our context and provide a simple method, \editsc, for translating component attributions into model edits.
We apply this approach to edit model behavior
on individual examples (\S\ref{subsec:examples}), classes (\S\ref{subsec:class_forget}), subpopulations
(\S\ref{subsec:subpops}), and concepts (\S\ref{subsec:backdoor}, \S\ref{subsec:typographic}).
Our findings indicate that \methodsc \emph{directly} enables model editing.

\paragraph{Problem setup.}
Consider a machine learning model $M$,
a target distribution over examples $\mathcal{D}_T$,
and a reference distribution $\mathcal{D}_R$.
A {\em model edit} on $M$ is an intervention that aims to modify performance
on the target examples $z \sim \mathcal{D}_T$ in a specific way,
while leaving behavior on reference examples $z \sim \mathcal{D}_R$ unchanged.
In its most general version,
model editing can involve additional training (e.g., constrained fine-tuning
\citep{zhu2020modifying} or rank-one parameter updates \citep{bau2020rewriting}),
targeted modifications to model parameters (e.g., weight pruning \citep{de2021sparse} or hypernetworks \citep{mitchell2021fast}),
or even architectural modifications (e.g., adaptors~\cite{hartvigsen2022aging} or adding neurons \citep{huang2023transformer})

Since our goal is to study model predictions in terms of model components,
we restrict model edits to {\em ablation-based} interventions in this work.
That is,
we only consider interventions whose output can be expressed
as component counterfactuals (see \Cref{eq:ablation}).
The goal of an editing method, then, is to identify a subset of model components whose ablation changes performance on a given target
set of examples $S_T$,
without impacting model behavior
on a reference set of examples $S_R$.
Definition \ref{def:edit} turns this intuition into a precise definition of the
ablation-based model editing problem.

\begin{definition}[Editing models by ablating components]
    \label{def:edit}
    Consider a model $M$ with computation graph $G_M$, component set $C = \{c_1,\ldots,c_N\}$, and model output function $f_M$.
    Let $\mathcal{D}_{\text{T}}$ and $\mathcal{D}_{\text{R}}$ denote target
    and reference distributions over examples, respectively.
    An $(\epsilon, \delta)$-{{effective} model edit} for
    $\mathcal{D}_{\text{R}}$ and $\mathcal{D}_{\text{T}}$ is
    an intervention that ablates a subset of components $C_{\text{edit}} \in 2^C$ such that
    \begin{align}
        \label{eq:effective}
        \underbrace{
        \mathbb{E}_{\mathcal{D}_{\text{R}}}[|f_M(z, C_{\text{edit}})-f_M(z, \emptyset)|]
        }_{\text{Effect of edit on reference examples is small}}
        \leq \epsilon
        \qquad \text{ and } \qquad
        \underbrace{
        \mathbb{E}_{\mathcal{D}_{\text{T}}}[|f_M(z, C_{\text{edit}})-f_M(z, \emptyset)|]
        }_{\text{Effect of edit on target examples is large}}
        \geq
        \delta,
    \end{align}
    where $f_M(z, C)$ denotes the component counterfactual function~\eqref{eq:ablation}, i.e., model output function $f_M$ evaluated on example $z$ after ablating components $C$.
\end{definition}
\noindent
As per \Cref{def:edit}, each component subset $C' \subset 2^C$ defines a potential model edit.
That is, effectively editing the model (as in \eqref{eq:effective}) requires
identifying a subset $C'$ that, when ablated, significantly changes
model outputs on the target distribution but not on the reference distribution.
A naive approach to this task would thus require searching over the
(combinatorial) space of all possible component subsets.
Is there a better way?

\paragraph{An attribution-based approach to model editing.}
In this section, our goal is to show that {an effective} component
attribution method (such as \methodsc) can directly serve as a guide
for identifying {effective} model edits.
Key to this utility is a fundamental connection between the attribution problem and the editing problem.
In particular, the former answers questions of the form,
``{\em how would the model outputs change if we were to ablate a subset of components?}''
while the latter inverts this question to ask
``{\em which components, when ablated, would change model outputs in a specific way?}''
By {identifying} the model components that are most ``important'' to the desired model outputs, an attribution method can thus identify a subset of model components to {target} through ablation-based editing (see \Cref{def:edit}).

To make this concrete, we propose \editsc, a simple three-step editing approach based on \methodsc attributions.
Specifically, given a model $M$ with a set of model components $C$, {a set of} target examples $S_T$ sampled
from $\mathcal{D}_T$, and {a set of} reference examples $S_R$
{sampled} from $\mathcal{D}_R$,
\editsc identifies a \skipthis{effective} {model} edit \eqref{eq:effective} in three steps:
\begin{enumerate}
\item Estimate \methodsc attributions $\smash{\bm{\theta}^{(z)} \coloneqq (\bm{w}^{(z)}, b^{(z)})}$ where $\bm{w}^{(z)} \in \mathbb{R}^{|C|}$ and $b^{(z)} \in \mathbb{R}$ for every target and reference example $z \in S_T \cup S_R$.
\item For each model component $c_i \in C$, use a simple $t$-test in order to quantify the ``importance'' of component $c_i$ to set of target examples $S_T$ relative to the set of reference examples $S_R$:
\begin{align}
    \label{eq:score}
    \tau(c_i) \coloneqq \frac{\mu_i(S_T)-\mu_i(S_R)}{\sqrt{\frac{\sigma_i^2(S_T)}{|S_T|}+\frac{\sigma_i^2(S_R)}{|S_R|}}}, \qquad \text{ where } \quad \begin{cases}
        \mu_i(S) = \frac{1}{|S|}\sum_{z \in S} {w}^{(z)}_i \\
        \sigma_i^2(S) = \frac{1}{|S|}\sum_{z \in S} ({w}^{(z)}_i - \mu(S))^2.
    \end{cases}
\end{align}
\item To increase model outputs on target examples, ablate a set of components $C_{\text{edit}}$ that contains the $k$ most negative
scores $\tau_i$, i.e., set
\begin{equation}
    \label{eq:c-edit}
    C_{\text{edit}} = \arg\text{bottom-}k(\{\tau(c_i): c_i \in C\}),
\end{equation}
where the number of ablated components $k$ is a hyperparameter which one can set, e.g., by cross-validation.
Similarly, if the goal is to {\em decrease} model outputs on $S_T$, we replace $\text{bottom-}k$ with $\text{top-}k$ in \eqref{eq:c-edit}.
\end{enumerate}
To make sense of the approach above,
note that for every component $c_i$ and a set of examples $S$,
$\mu(S)$ in \Cref{eq:score} leverages attributions to directly estimate $\mu_i(S)$, the average effect of ablating $c_i$ on model predictions for samples in set $S$.
Similarly, the term $\sigma_i^2(S)$ captures the variation (across examples) of this effect.
As a result, the score $\tau(c_i)$ in \Cref{eq:score} exactly corresponds to the two-sample $t$-test statistic, with a null hypothesis that the  component $c_i$ has an equal average effect on the target distribution $\mathcal{D}_T$ and the reference distribution $\mathcal{D}_R$.
We then use these scores $\{\tau(c_i) : {c_i \in C}\}$ in \Cref{eq:c-edit} in order to identify components that, when ablated, would change outputs on target examples the most relative to the change in the outputs of reference examples.

\begin{remark}[Ablation-based edits]
    \label{rem:coarse}
    Our restriction of model edits to ablations of component subsets in \editsc is a significant one.
    In particular, simply ablating model components is likely \underline{\smash{not}}
    the most effective way of editing a model.
    Our goal in this section, however, is not to propose a state-of-the-art model editing
    method, but rather to answer two questions.
    First, we want to assess the practical utility of
    \methodsc for informing model edits.
    Second, we want to shed light on whether
    large-scale models are amenable to zeroth-order editing, i.e., without gradient information.
    Leveraging \methodsc attributions in conjunction with editing techniques based on fine-tuning in order to make finer-grained model edits is an interesting direction for future work.
    \end{remark}

Next, we use \textsc{Coar-Edit} to edit large-scale models
trained on classification tasks, where we use correct-class
margin~\eqref{eq:margin} as the model output function.
Specifically, we conduct five experiments to evaluate the effectiveness of \editsc in editing model behavior using only a few examples and without requiring additional training.
\begin{itemize}
    \item[(a)] In \Cref{subsec:examples}, we correct individual model errors
    without impacting overall performance;
    \item[(b)] In \Cref{subsec:class_forget}, we selectively ``forget'' a
    specific class while preserving model performance on other classes;
    \item[(c)] In \Cref{subsec:subpops}, we start with a model that performs
    disparately across a set of subpopulations, and edit the model
    to improve its accuracy on underperforming subpopulations;
    \item[(d)] In \Cref{subsec:backdoor}, we localize (known)
    backdoor attacks and mitigate them by ablating a small number of components;
    \item[(e)] In \Cref{subsec:typographic}, we edit
    CLIP classifiers to be more robust to typographic attacks.
\end{itemize}
For all experiments, we provide additional details and analyses in \Cref{app:additional_editing}.

\clearpage
\subsection{Editing individual model predictions}
\label{subsec:examples}
In this section, we test whether \editsc can
modify individual predictions of an ImageNet ResNet50 classifier (Setup B in \Cref{sec:eval}) without impacting its overall performance.
Specifically, we study the case where the target distribution $\mathcal{D}_T$ is a singleton
example on which we want to improve performance.
An effective model edit in this context (\Cref{def:edit}) would increase the model's
margin \eqref{eq:margin} on $z$ to be greater than zero without affecting aggregate model performance.

\begin{figure}[!t]
    \centering
    \includegraphics[width=\textwidth]{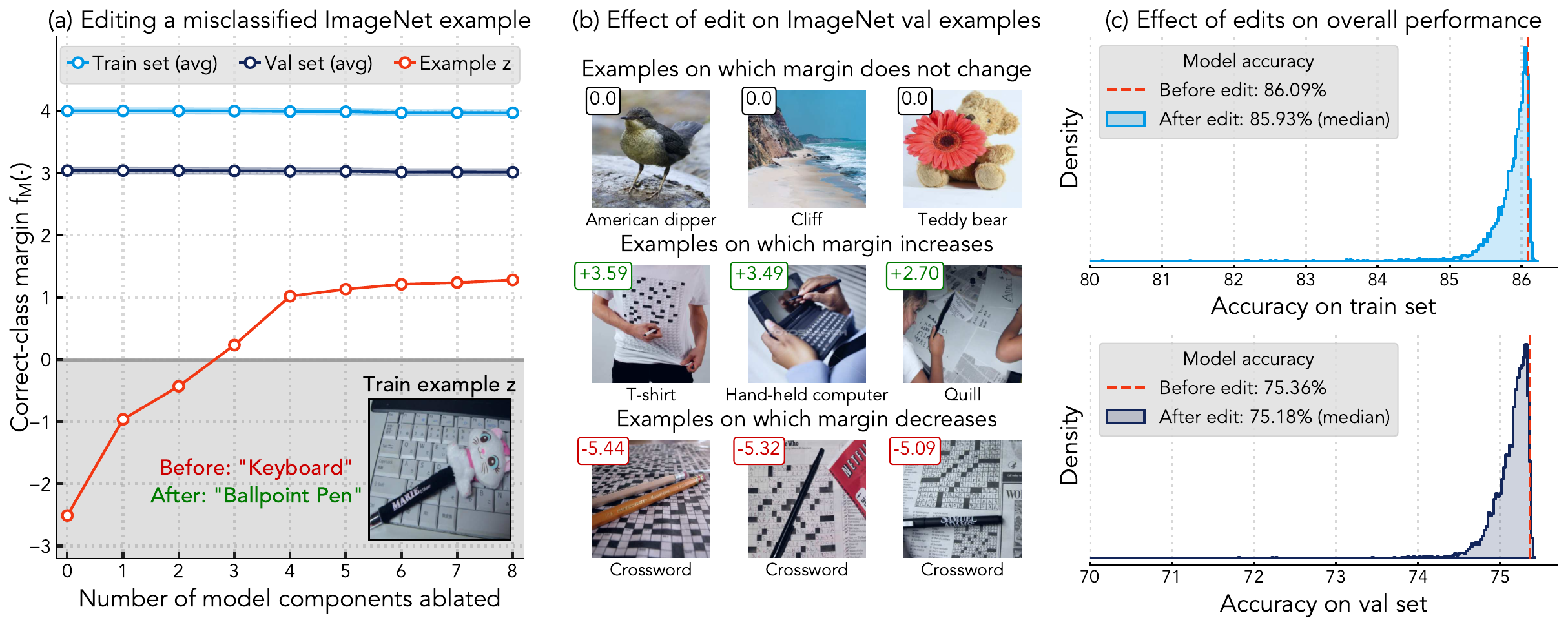}
    \caption{
    \textbf{Editing individual model predictions with \editsc.}
    We edit a ResNet50 model to correct a misclassified ImageNet example $z$ shown on the left.
    Specifically, ablating a few components via \editsc (see \eqref{eq:c-edit}) increases the correct-class margin \eqref{eq:margin} on example $z$ (red) without changing the average margin on the train set (light blue) or validation set (dark blue).
    In the center panel, we observe that the examples on which model outputs change the least (top row) due to the edit are visually dissimilar to example $z$ as well as examples on which model outputs change most positively (middle row) and negatively (bottom row).
    On the right, we find that individually performing model edits to correct {every} misclassified example in the validation set
    incurs a median accuracy drop of at most $0.2\%$ on the train set (top row) and validation set (bottom row).
    }
    \label{fig:example}
\end{figure}

\paragraph{Results.} We apply {\editsc}
to edit individual misclassified examples $z$, setting $S_T = \{z\}$
and $S_R$ to be a small set of random samples from the ImageNet dataset.
We present our findings in~\Cref{fig:example}.
\Cref{fig:example}a illustrates a single such edit,
where we correct the model's prediction on a specific ImageNet example
from ``keyboard'' to ``ballpoint pen'' by ablating $k = 3$ components
($0.01\%$ of all components).
Specifically, increasing the number of ablated components $k$ consistently improves the
correct-class margin on target example $z$ (red) without changing the average margin over the training set (light blue) or validation set (dark blue).
\Cref{fig:example}b then visualizes  (again, for the specific example being edited in \Cref{fig:example}a) the examples on which model outputs changes most (and least) drastically.
{A} qualitative inspection {here} suggests that examples with unchanged margins are
dissimilar to $z$ (first row), whereas examples that are most
positively (second row) or negatively (third row) impacted share similar visual
features with $z$, {e.g., pen-like objects.}
Finally, \Cref{fig:example}c shows that we can individually fix \emph{every}
misclassification in the ImageNet validation set while incurring a median accuracy drop
of $0.2\%$ on the training set (top row) and validation set (bottom
row).
We defer experiment details and additional results to~\Cref{app:example_level}.

\begin{figure}[!t]
    \centering
    \includegraphics[width=\textwidth]{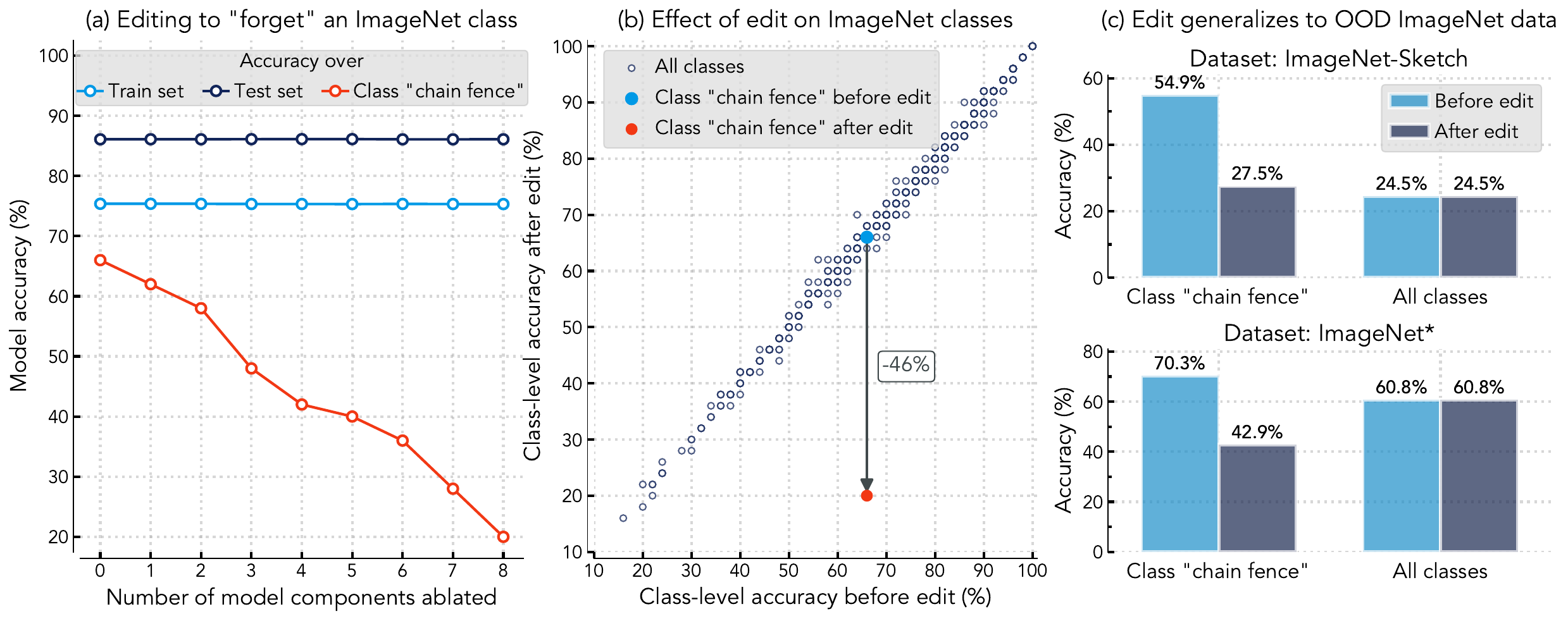}
    \caption{
    \textbf{``Forgetting'' a class with \textsc{Coar-Edit}.} We edit an ImageNet-trained ResNet-50
    (Setup B from \Cref{sec:eval}) to selectively degrade performance on the ``chain-link fence'' class.
    On the left, we observe that increasing the number of components $k$ ablated via \editsc decreases model accuracy on the ``chain-link fence'' class (red) while preserving overall accuracy on the train and validation set.
    In the center panel, we compare class-wise accuracies before and after performing the model edit %
    and observe a significant accuracy drop on the ``chain-link fence'' class but not on other classes.
    On the right, we find that the edit transfers to distribution-shifted versions of ImageNet (ImageNet-Sketch \cite{wang2019learning} and ImageNet$\star$ \cite{vendrow2023dataset}) as targeted, i.e., degrading performance on the ``chain-link fence'' class without changing average performance. %
    }
    \label{fig:class}
\end{figure}

\subsection{``Forgetting'' a class}
\label{subsec:class_forget}
We now consider ``selective forgetting'' problem \cite{wang2023comprehensive},
where the goal is to impair model performance on (only) a specific set of examples.
In this experiment, we edit the same ImageNet ResNet-50 model
(Setup B) as in \Cref{subsec:examples},
with the goal of forgetting the {entire} ``chain-link fence'' class.
Like before,
we use our editing approach \textsc{Coar-Edit} (see \eqref{eq:score} and \eqref{eq:c-edit})
to identify components that, when ablated, decrease the model's correct-class margin on
examples from the ``chain-link fence'' class, but not on reference examples from
other classes.

\paragraph{Results.} \Cref{fig:class} summarizes our findings.
In~\Cref{fig:class}a, we show that ablating just eight (out of $22,720$) model
components degrades accuracy on the ``chain fence'' class from $66\%$ to $20\%$
while preserving overall accuracy on the train and validation set.
Then, in~\Cref{fig:class}b, a comparison of class-wise accuracies before and
after the edit shows that our approach specifically targets the ``chain fence''
class without impacting performance on any other class.
Finally, in~\Cref{fig:class}c, we evaluate model performance on ImageNet-Sketch \citep{wang2019learning} (top)
and ImageNet$\star$ \citep{vendrow2023dataset} (bottom) datasets to show that the our edit is
robust to distribution shifts in both the target and reference distribution.
Through additional experiments in~\Cref{app:class_level}, we highlight that
(a) our approach is sample-efficient, not needing many samples from the target
and reference distributions to find effective edits; and
(b) our findings are robust to the choice
of class to forget.

\subsection{Improving subpopulation robustness}
\label{subsec:subpops}

Machine learning models often latch onto spurious correlations in the training
dataset \cite{geirhos2019imagenet,shah2020pitfalls,hermann2023foundations},
{resulting in subpar} performance on subpopulations where these correlations do
not hold~\cite{buolamwini2018gender,oakden2020hidden}.
In this section, we test whether our editing approach can boost performance on
such underperforming subpopulations without degrading overall performance.

In particular, we evaluate \textsc{Coar-Edit} on two benchmark
datasets---Waterbirds \cite{sagawa2020distributionally} and
CelebA \cite{liu2015deep}---where models fare poorly on
subpopulations that are underrepresented in the training data.
On both datasets, our goal is to improve a given model's
{\em worst-{subpopulation} accuracy}---we defer experiment details to \Cref{app:group_level}.

\begin{figure}[!t]
    \centering
    \includegraphics[width=\textwidth]{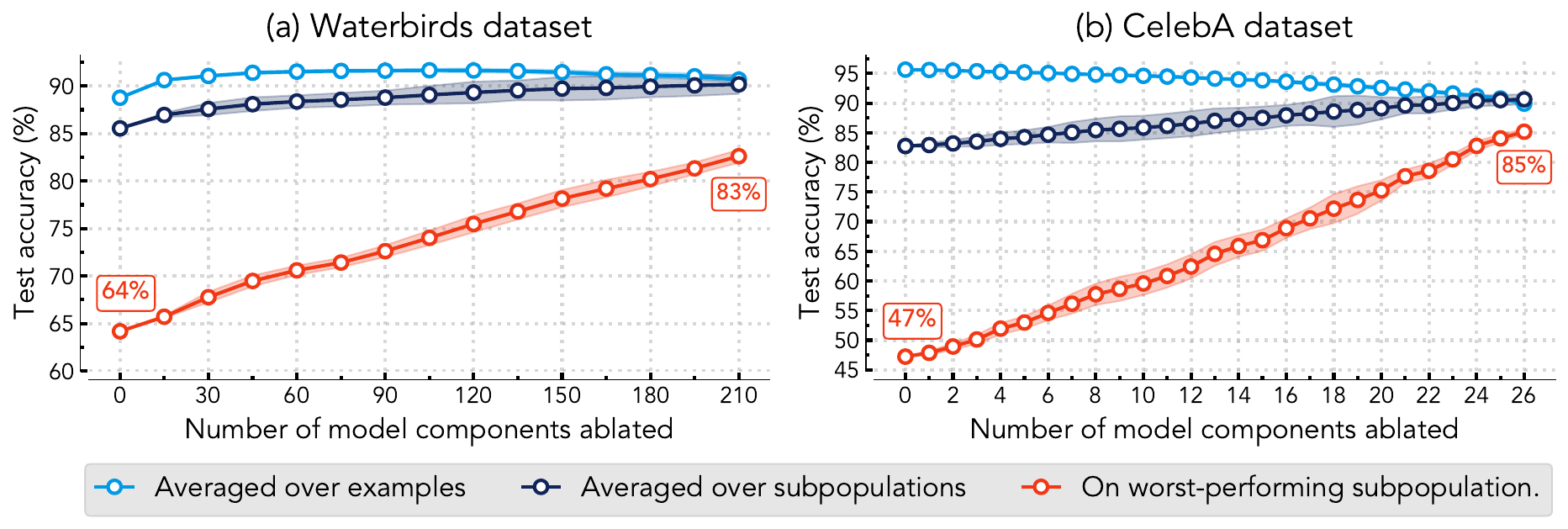}
    \caption{\textbf{Improving subpopulation robustness with \textsc{Coar-Edit}.}
    We edit pre-trained ResNet-50 models to improve their worst-subpopulation accuracy on two benchmark datasets: Waterbirds \cite{sagawa2020distributionally} and CelebA \cite{liu2015deep}.
    Before applying \editsc, models fine-tuned on Waterbirds and CelebA attain $87\%$ and $96\%$ test accuracy but only $64\%$ and $47\%$ accuracy on their worst-performing subpopulations, respectively.
    On the left, applying \editsc by ablating $210$ of $22,720$ components in the Waterbirds model increases worst-subpopulation accuracy from $64\%$ to $83\%$ without degrading its accuracy averaged over examples (light blue) and subpopulations (dark blue).
    Similarly, on the right, editing the CelebA model by ablating a targeted subset of $26$ components improves worst-subpopulation accuracy from $47\%$ to $85\%$. %
    }
    \label{fig:groups}
\end{figure}

\paragraph{Results.} On both datasets, \methodsc successfully identifies
component subsets that correspond to {effective} model edits.
\Cref{fig:groups} depicts our results.
On Waterbirds (\Cref{fig:groups}a),
ablating $210$ components ($0.9\%$ of all components)
improves {worst-subpopulation accuracy}
from $64\%$ to $83\%$ (red) without degrading its accuracy
averaged uniformly over examples (light blue) and subpopulations (dark blue).
On CelebA, \Cref{fig:groups}b demonstrates that zeroing out $26$ of
$22,720$ model components improves
{worst-subpopulation accuracy}
from $47\%$ to $85\%$
and {average-subpopulation accuracy}
from $84\%$ to $90\%$
while only incurring a $5\%$ drop in test set accuracy.

Before continuing, we make two observations.
First, on both datasets, our editing-based approach is
\emph{sample-efficient}---it does not require subpopulation-level annotations for the training set,
and only uses $20$ random {training} examples from each subpopulation to find effective model edits.
Second, our results indicate that simply ablating a few components from models trained
via ``standard'' empirical risk minimization {(ERM)} can achieve worst-subpopulation
accuracy improvements that are comparable to gains from specialized methods
(e.g., based on robust optimization~\cite{sagawa2020distributionally}, dataset balancing~\cite{idrissi2022simple}, and generative modeling~\cite{goel2020model}).

\subsection{Localizing backdoor attacks}
\label{subsec:backdoor}
\begin{figure}[!t]
    \centering
    \includegraphics[width=\textwidth]{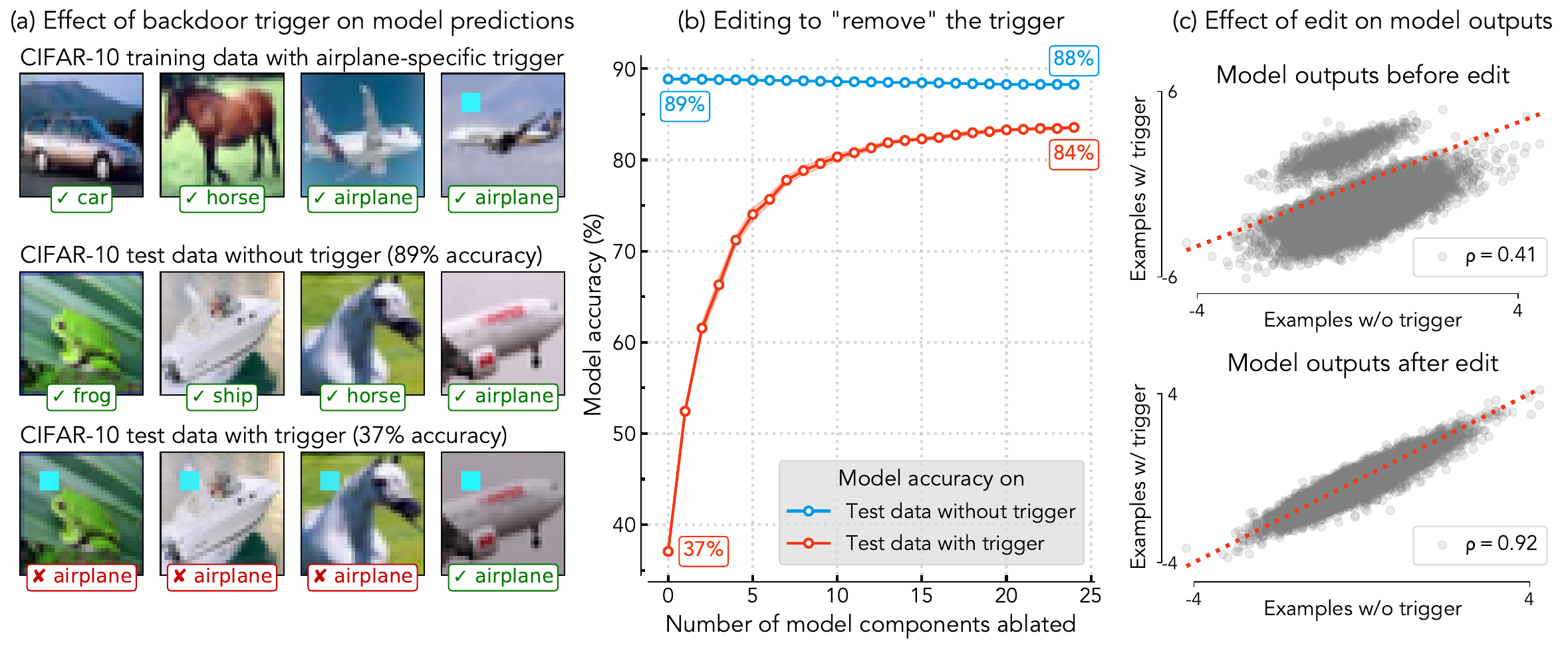}
    \caption{
        \textbf{Localizing backdoor attacks with \editsc.}
        We edit a ResNet18 model trained on a backdoored CIFAR-10 dataset in which half of all training examples in the ``airplane'' class contain a planted blue-squared trigger.
        On the left, we find that the model is sensitive to the trigger---backdoor attacks that add the trigger to examples drop test accuracy from $89\%$ (middle row) to $37\%$ (bottom row).
        In the center panel, we apply \editsc to identify $25$ backdoor-specific components that, when ablated, boost accuracy on examples with the trigger (red) from $37\%$ to $84\%$ without impacting accuracy on examples without the trigger (blue).
        On the right, we find that the edit suppresses sensitivity to the trigger even at the example level---the correlation between model outputs on paired examples with and without the trigger increases from $0.41$ to $0.92$.}
    \label{fig:concept_cifar}
\end{figure}

We now use \editsc to analyze the sensitivity of model predictions to backdoor attacks~\cite{biggio2012poisoning,gu2017badnets}, where an adversary plants a spurious correlation in the training dataset and uses it as a trigger to override predictions at test time.
In this experiment, we first train a ResNet18 model on a modified CIFAR-10 dataset in which half of the training examples in the ``airplane'' class contain a planted blue-squared trigger, as shown in~\Cref{fig:concept_cifar}a.
Then, using \editsc, we evaluate whether the effect of this trigger on predictions can be localized to a few components, which, if ablated, induces model robustness to backdoor attacks without degrading overall performance.

\paragraph{Results.}
\Cref{fig:concept_cifar} summarizes our findings.
\Cref{fig:concept_cifar}a  shows that prior to editing, the model trained on the modified CIFAR-10 dataset (top row) is sensitive to backdoor attacks---simply adding the ``airplane'' trigger to test examples drops model accuracy from $89\%$ (middle row) to $37\%$ (bottom row).
To localize the effect of the trigger, we use \editsc over ten \emph{paired} examples---i.e., examples with and without the backdoor trigger---to identify a few components that, when ablated, correct the misclassifications induced by the trigger without impacting predictions on test examples without the trigger.
In \Cref{fig:concept_cifar}b, we find that editing the model by ablating $25$ components ($1\%$) is sufficient to boost accuracy on test examples with the trigger (red) from $37\%$ to $84\%$ without impacting accuracy on examples without the trigger (blue) by more than $1\%$. %
\Cref{fig:concept_cifar}c shows that the edit suppresses the effect of the trigger at the example level as well, improving correlation between model outputs on examples with and without the trigger from $0.41$ (top row) to $0.92$ (bottom row).
We defer additional details and analyses to~\Cref{app:backdoor}.

\begin{figure}[!t]
    \centering
    \includegraphics[width=\textwidth]{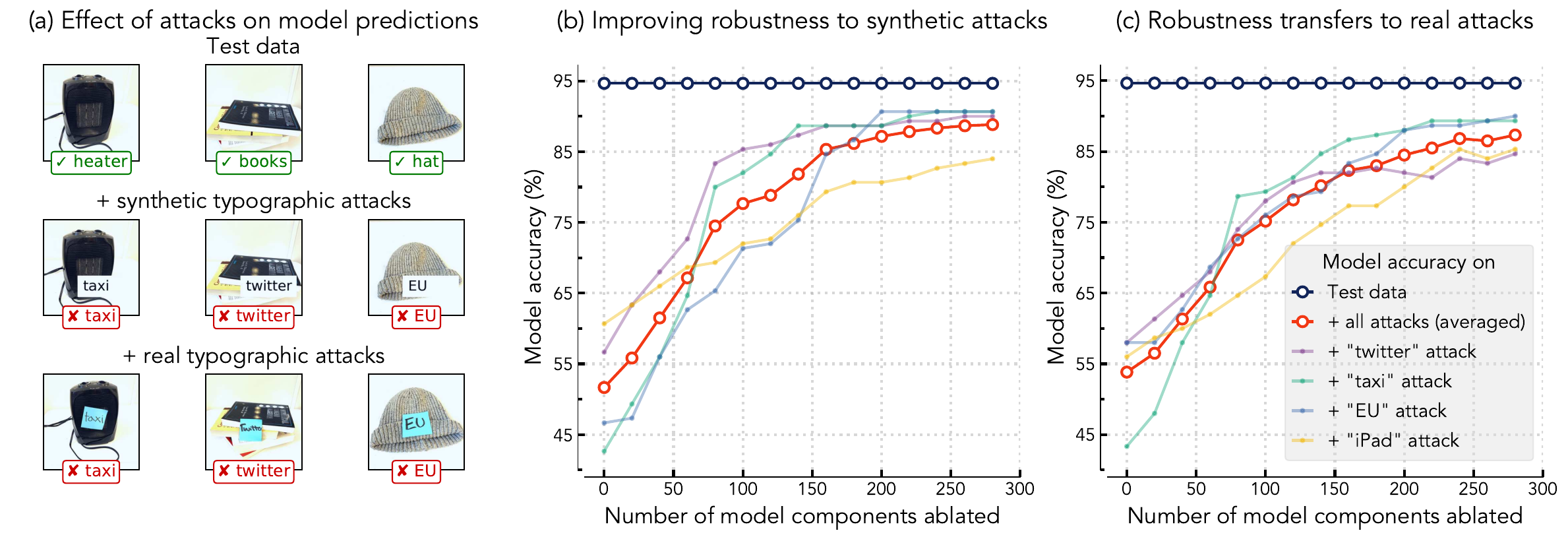}
    \caption{
        \textbf{Improving robustness to typographic attacks with \editsc.}
        We edit a zero-shot CLIP ViT-B/16 classifier to improve its robustness to typographic attacks~\cite{goh2021multimodal}.
        On the left, we find that predictions on images of household objects (top row) can be manipulated to ``taxi'', ``twitter'', or ``EU'' via synthetic (middle row) and real (last row) typographic attacks.
        In the center panel, we use \editsc to identify components that, when ablated, improve average accuracy on examples with {synthetic} typographic attacks (red) from $51\%$ to $89\%$ while maintaining accuracy on examples without attacks (blue).
        Similarly, on the right, we find that the edit transfers robustness to real typographic attacks as well, improving average accuracy from $54\%$ to $86\%$.}
    \label{fig:concept_clip}
\end{figure}

\subsection{Improving robustness to typographic attacks}
\label{subsec:typographic}

Zero-shot CLIP classifiers are vulnerable to typographic attacks~\cite{goh2021multimodal} that simply insert text to images in order to induce misclassifications.
In this experiment, we evaluate whether our editing approach can improve the robustness of a zero-shot CLIP ViT-B/16 classifier using a dataset~\cite{materzynska2022disentangling} comprising $180$ images with and without multiple typographic attacks.
Specifically, we use \editsc to identify a subset of components that, when ablated, correct the misclassifications induced by synthetic typographic attacks without impacting predictions on images without attacks.
We defer additional details to~\Cref{app:typographic}.

\paragraph{Results.}
\Cref{fig:concept_clip} summarizes our findings.
In~\Cref{fig:concept_clip}a, we show that the predictions of a zero-shot CLIP ViT-B/16 classifier on images of household objects (top row) can be manipulated to ``taxi'', ``twitter'', or ``EU'' via synthetic (middle row) or real (last row) typographic attacks.
More quantitatively, we find that the zero-shot accuracy on images with synthetic and real typographic attacks drops from $95\%$ to $51\%$ and $54\%$, respectively.
\Cref{fig:concept_clip}b shows that ablating a subset of $300$ components ($0.4\%$) identified via \editsc improves the accuracy on held-out images with synthetic typographic attacks from $51\%$ to $89\%$ on average (red), without impacting accuracy on images without attacks (dark blue).
Furthermore, in~\Cref{fig:concept_clip}c, we find that our edit transfers robustness to real typographic attacks as well, improving accuracy on held-out images from $54\%$ to $86\%$ on average.
Similar to previous experiments, our approach is sample-efficient in that it only requires $15$ pairs of target and reference examples with and without synthetic attacks to identify the edit described above.

To summarize, simply ablating targeted subsets of components identified via
\editsc can induce specific model behavior without requiring additional
training. %
More broadly, our findings highlight how accurate component attribution
\emph{alone} can directly inform model editing.

\section{Related work}
\label{sec:related}
Our work relates to multiple topics in interpretability, which we categorize into works that focus on localizing model behavior, interpreting individual components, editing models, and approximating functional behavior via interpretable proxies.

\medskip\noindent\textbf{Localizing model behavior.}
One line of work in mechanistic interpretability
attempts to identify ``circuits'' \citep{olah2020zoom} or ``subnetworks''
\citep{cao2021low} within neural networks that are responsible for specific
capabilities or biases such as factual recall \cite{meng2022locating}, gender
bias \cite{vig2020investigating}, or arithmetic reasoning
\cite{stolfo2023understanding}.
Building on this, several works focus on automated localization of behavior
using techniques such as fine-tuning~\cite{panigrahi2023task}, activation
patching \cite{conmy2023towards,goldowsky2023localizing}, or differentiable
masking \cite{bayazit2023discovering,de2021sparse,chang2023localization}.
More recent studies evaluate these methods in terms of their sensitivity to
design choices \cite{zhang2023towards}, usefulness for model editing
\cite{hase2023does}, and ability to characterize functional behavior
\cite{wen2023transformers}.
Other works develop metrics to quantify the ``importance'' of individual
components (e.g.,
\citep{dhamdhere2018important,leino2018influence,ghorbani2020neuron}), which we
adapt as baselines in~\Cref{sec:eval}.
Rather than localizing human-defined behavior to specific components, the
component modeling task (\Cref{def:attribution}) aims to explicitly model the
collective function mapping component ablations to predictions.

\medskip\noindent\textbf{Interpreting specific model components.}
The works discussed above aim to localize specific model behavior to
components; another line of work studies the reverse direction, and
introduces methods for uncovering the functionality corresponding to a specific model
component.
Such methods include feature
visualization \cite{zeiler2014visualizing,ghiasi2022vision},
activation maps \cite{bau2017network,mu2020compositional},
ablations \cite{zhou2018revisiting}, saliency maps~\cite{olah2018building},
probing \cite{dalvi2019one,durrani2020analyzing}, and natural language
descriptions \cite{hernandez2021natural,oikarinen2022clip,bills2023language}.
Subsequent analyses use these methods to identify and
ascribe meaning to specific model components by labeling them as, e.g.,
``curve detectors'' \cite{cammarata2020curve}, ``knowledge
neurons'' \cite{dai2021knowledge}, ``multimodal
neurons'' \cite{goh2021multimodal}, or ``syntax
units'' \cite{lakretz2019emergence}.
More recent work revisits the reliability and robustness of such methods
\cite{geirhos2023dont,bolukbasi2021interpretability,hooker2018benchmark,shah2021input,hewitt2019designing,antverg2021pitfalls,huang2023rigorously}.
Here, our goal is not to interpret specific model components,
but rather to study how all components jointly influence model predictions
through the lens of component modeling (\Cref{def:problem}).

\medskip\noindent\textbf{Editing model behavior.}
Another related line of work focuses on {\em model editing}, the goal
of which is to make small, targeted changes to model representations
in order to induce or suppress a specific behavior.
Model editing methods include \skipthis{using}
``hypernetworks''~\cite{de2021editing,mitchell2021fast},
\skipthis{making} rank-one updates to
model parameters~\cite{bau2020rewriting,santurkar2021editing,meng2022locating},
constrained fine-tuning \citep{zhu2020modifying},
and weight interpolation \cite{ilharco2022editing,zou2023representation},
among other methods.
Recent work has also studied erasing concepts and suppressing spurious
correlations from models using layer-wise linear
probing~\cite{ravfogel2022linear}, CLIP-specific \skipthis{text-based} methods~\cite{gandelsman2023interpreting,chen2023interpreting}, and
fine-tuning variants~\cite{gandikota2023erasing,kirichenko2022last}.
In this work, we introduce \editsc to show that effective component attribution can directly enable model editing at the level of examples (\S\ref{subsec:examples}), classes (\S\ref{subsec:class_forget}), subpopulations (\S\ref{subsec:subpops}), and spurious concepts (\S\ref{subsec:backdoor}, \S\ref{subsec:typographic}) by zeroing out targeted subsets of components.

\medskip\noindent\textbf{Understanding machine learning models by proxy.}
More generally, our approach connects to a line of research that
aims to understand machine learning models by constructing
{\em interpretable proxies}.
For example, feature attribution methods like LIME \citep{ribeiro2016}
approximate a given ML model with a linear model in input space.
Similarly, datamodeling~\citep{ilyas2022datamodels}
approximates a given learning algorithm by a linear model
in ``dataset space.''
Another line of work analyzes the behavior of deep networks via high-level causal abstractions \citep{geiger2021causal,geiger2023causal}, user-specified causal graphs over task-specific variables.

\section{Discussion}
\label{sec:discussion}
In this section, we discuss connections between component attribution and model editing, discuss directions for future work, and outline key limitations of \methodsc.

\paragraph{Does localization help with model editing?}
The extent to which localizing specific model behavior to a subset of model components helps with model editing remains contested. %
On one hand, \citet{hase2023does} show that localizing factual associations in language models does not necessarily help with editing these associations.
More broadly, recent evaluation studies suggest that model editing can degrade robustness to distribution shifts~\cite{brown2023robustness} and may not modify model behavior in a consistent manner~\cite{cohen2023evaluating}.
On the other hand, recent work shows that some localization methods can in fact recover ground-truth localization in controlled settings~\cite{chang2023localization} and improve calibration of fine-tuned language models~\cite{panigrahi2023task}.
Our findings in \Cref{sec:applications} substantiate the latter view, as \editsc uses component attributions to identify a target subset of components that, when ablated, modify model behavior as intended.

\paragraph{Future work.}
We highlight three directions that,
while outside the scope of this work, may be interesting avenues for future work. %
\begin{itemize}[leftmargin=*]
    \item \textbf{Analyzing neural network representations.} An {interesting} direction for future work could be
    to use component attribution (and component models, more generally)
    to study empirically documented phenomena in deep learning.
    There are a plethora of questions to ask here which,
    although beyond the scope of this work,
    are natural applications of component attributions.
    For example, extending our results from \Cref{subsec:examples},
    can we use component attribution to better isolate
    ``opposing signals'' \citep{rosenfeld2023outliers} for a given task, and to understand their role in shaping model predictions?
    Can we use component attributions to study how model predictions change due to adversarial perturbations~\citep{goodfellow2015explaining}, or over the course of training~\citep{kalimeris2019sgd}?
    Similarly, can we develop improved methods for localizing memorized inputs to specific model components \citep{feldman2020neural,maini2023can}?
    Given that component attributions are causally meaningful,
    can we use them as a kernel with which to compare different models~\citep{kornblith2019similarity} or learning algorithms~\citep{shah2023modeldiff}?
    \item \textbf{Attributing generative models.}
    While we focus on vision models in this work, \methodsc is a general method that can estimate component attributions for any machine learning model.
    Future work might thus explore possible
    model output functions (and their corresponding component attributions)
    for {\em generative} models.
    For diffusion-based generative models,
    one might study the denoising error for a fixed timestep, as in \citep{georgiev2023journey,zheng2023intriguing}.
    For language models, a possible point of start (following \citet{park2023trak}) would be to use the average correct-class margin \eqref{eq:margin} of a sequence of tokens as the model output function.
    \addlm{Our preliminary experiments in~\Cref{app:language_models} show that \methodsc learns accurate component attributions for language models such as GPT-2~\cite{radford2019language} and Phi-2~\cite{li2023textbooks}.}
    In general, estimating and applying component attributions for generative models is a promising avenue for future work.
    \item \textbf{Beyond linear component attribution.}
    The fact that component attributions' predictiveness decreases on
    out-of-distribution component subsets, i.e., when $\alpha_\text{test} \neq
    \alpha_\text{train}$, suggests that the linear form of
    component attributions might not be
    expressive enough to fully capture the map between model components and
    outputs.
    Given the generality of \methodsc, an interesting avenue for future work would be to explore whether non-linear component models such as decision trees or kernel methods  predict component counterfactuals more accurately, and as a result, improve model editing.
\end{itemize}

\paragraph{Limitations.}
Our attribution-based approach for decomposing and editing model predictions is not without its limitations.
First, estimating \methodsc attributions involves constructing datasets of ground-truth counterfactuals (\Cref{eq:dataset}), which can require a large number of forward passes through the model.
In \Cref{app:sample_complexity}, we show that simply using $2$-$5\times$ fewer samples can significantly speed up \methodsc without impacting the quality of the resulting attributions.
Mitigating this computational bottleneck further through better sampling or approximation techniques is an interesting avenue for future work.
Second, \methodsc requires specifying an ablation method (\Cref{eq:ablation}). While we use zero ablations in this work (\Cref{rem:ablation}), one could also use \methodsc with ablation methods (e.g.,~\citet{chan2022causal}) that account for distribution shifts in activation space.
For example, in \Cref{app:alpha_eval}, we show that \methodsc yields predictive attributions with another ablation method that simply scales down the activations of ablated components by a small constant factor instead of setting them to zero.
Finally, as noted in \Cref{rem:coarse}, our model editing approach \methodsc is coarse-grained in that it modifies model behavior by simply ablating a targeted subset of components.
Using \methodsc in conjunction with gradient-based editing techniques in order to make finer-grained model edits is an interesting avenue for future work.

\section{Conclusion}
\label{sec:conclusion}
We first formalize the problem of decomposing model predictions in terms of model components through the \emph{component modeling} task.
We specifically focus on a special case of component modeling, \emph{component attribution}, where the goal is to predict the counterfactual impact of every component on a given prediction. %
We then propose \methodsc, a scalable method for estimating predictive component attributions, and demonstrate its effectiveness across model architectures, datasets, and tasks.
Finally, {through a series of five experiments}, we also stress-test the utility of \methodsc attributions in directly editing model behavior without requiring additional training. %

\section*{Acknowledgements}
\label{sec:acknowledgements}
The authors would like to thank Benjamin Cohen-Wang, Logan Engstrom, Alaa Khaddaj, and Kristian Georgiev for helpful discussions and  comments on an earlier draft of this manuscript.

Work supported in part by the NSF grant DMS-2134108. This material is based upon work supported by the Defense Advanced Research Projects Agency (DARPA) under Contract No. HR001120C0015.

\clearpage
\printbibliography

\clearpage
\appendix
\addcontentsline{toc}{section}{Appendix} %
\renewcommand\ptctitle{Appendices}
\part{}
\parttoc

\clearpage
\section{Evaluation setup}
\label{app:additional_expt_details}
In this section, we outline the experiment setup---datasets, models, baselines, implementation details---used in \Cref{sec:eval} to evaluate whether \methodsc attributions can accurately estimate ground-truth component counterfactuals. %

\subsection{Pseudocode}
\label{app:pseudocode}

\begin{algorithm}
    \caption{An outline for estimating component attributions with \methodsc.}
    \begin{algorithmic}[1]
        \Procedure{Coar}{example $z$, model $M$ with output function $f_M$ and components $C$, ablation frac. $\alpha$}
        \State Set $D^{(z)} \gets []$ \Comment{\emph{initialize component dataset \eqref{eq:dataset}}}
        \For{$i \in \{1,\ldots,m\}$} \Comment{\emph{$m$ denotes dataset size}}
        \State Sample a subset $C_i \subset C$ from $\mathcal{D}_C$ where $|C_i| = \alpha \cdot |C|$
        \State Set $y_i \gets f_M(z, C_i)$ \Comment{\emph{compute component counterfactual \eqref{eq:ablation}}}
        \State Define $\bm{0}_{C_i} \in \{0, 1\}^|C|$ as $(\bm{0}_{C_i})_j = 0$ if $c_j \in C_i$ else $1$
        \State Update $D^{(z)} \gets D^{(z)} + [(\bm{0}_{C_i}, y_i)]$ \Comment{\emph{update component dataset}}
        \EndFor
        \State $\theta^{(z)}, b^{(z)} \gets$ \Call{LinearRegression}{$D^{(z)}$} \Comment{\emph{estimate component attributions via \Cref{eq:regression}}}
        \State\Return $\theta^{(z)}, b^{(z)}$
        \EndProcedure
    \end{algorithmic}
    \label{alg:pseudo}
\end{algorithm}

\subsection{Datasets and models.}
\label{app:datasets_and_models}

We now outline the datasets and models used to evaluate \methodsc (\S\ref{sec:eval}) and \editsc (\S\ref{sec:applications}). %

\paragraph{CIFAR-10.} We use the standard CIFAR-10~\citep{krizhevsky2009learning} image classification dataset to evaluate \methodsc attributions (\Cref{sec:eval}, \Cref{app:arch_eval}) and for an editing task (\Cref{subsec:backdoor}).
We train ResNet, ViT, and MLP models that attain test accuracies of $91\%$, $83\%$ and $56\%$ respectively.
We specify a computation graph over $2,344$ components for the ResNet-18 model, $31,728$ components for the ViT model, and $3,072$ components for the MLP model.
Each component in the ResNet-18 model corresponds to a convolution filter.
Similarly, each component in the ViT and MLP models corresponds to a neuron.

\paragraph*{ImageNet.} We use the standard ImageNet~\citep{deng2009imagenet} image classification dataset to evaluate \methodsc attributions in \Cref{sec:eval} and for editing tasks in \Cref{subsec:class_forget}.
We use ImageNet-Sketch \citep{wang2019learning} and five random shifts from  ImageNet$\star$ \citep{vendrow2023dataset}---{``in the water''}, {``at dusk simple''}, {``orange''}, {``pencil sketch''}, {``green''}---
to evaluate the out-of-distribution performance of edited ImageNet models in \Cref{subsec:class_forget}.
We use the pre-trained ResNet50 and ViT-B/16 models\footnote{Model and pre-trained weights taken from torchvision: \url{https://pytorch.org/vision/stable/models.html}} that attain test accuracies of $75.4\%$ and $80.7\%$ respectively.
For the ResNet-50 model, we specify a computation graph over $22,720$ components, each corresponding to a convolution filter.
Similarly, for the ViT-B/16 model, we specify a computation graph over $82,944$ components, each corresponding to a neuron.

\paragraph{Waterbirds.}
The Waterbirds dataset~\cite{sagawa2020distributionally} comprises images of birds taken from the CUB dataset \cite{wah2011caltech} and pasted on backgrounds from the Places dataset~\cite{zhou2017places}.
The task here is to classify ``waterbirds'' and ``landbirds'' in the presence of spurious correlated ``land'' and ``water'' backgrounds in the training dataset.
\citet{sagawa2020distributionally} introduce Waterbirds as a benchmark to improve model performance under subpopulation shifts induced by spurious correlations.
We use this dataset to evaluate whether \editsc can improve subpopulation robustness via model editing.
In this experiment, we fine-tune an ImageNet ResNet50 model and use a computation graph over $22,720$ components, each corresponding to a convolution filter.

\paragraph{CelebA.}
The CelebA dataset~\cite{li2020celeb} comprises images of celebrities with binary attributes such as ``smiling'', ``wearing hat'', ``wearing lipstick'', etc.
Similar to previous work on subpopulation robustness (e.g., ~\cite{sagawa2020distributionally}), we repurpose CelebA as a binary classification task where the goal is to predict whether a person in a given image has blond hair.
The attributes ``hair color'' and ``gender'' are spuriously correlated in the training dataset, resulting in models that latch on to a ``gender $\rightarrow$ blond hair'' shortcut and  underperform on the ``blond males'' subpopulation.
Similar to the Waterbirds setting, we fine-tune an ImageNet ResNet50 model and specify a computation graph over $22,720$ components, each corresponding to a convolution filter.

\paragraph{Typographic attacks dataset.}
We use a dataset of typographic attacks~\cite{materzynska2022disentangling} for an editing task in~\Cref{subsec:typographic}. This dataset comprises $180$ images of household objects with and without eight typographic attacks such as ``taxi'', ``twitter'', ``EU'', and ``iPad''. We visualize some examples from this dataset in~\Cref{fig:concept_clip}.
Our experiment in~\Cref{subsec:typographic} uses this dataset along with a zero-shot CLIP ViT-B/16 classifier~\cite{radford2021learning}.
For this model, we specify a computation graph over all $82,944$ components, corresponding to the set of all weight vectors (individual rows in weight matrices) in all self-attention and MLP modules.
See~\Cref{app:typographic} for more details.

\paragraph*{TinyStories.}
We use the TinyStories dataset~\cite{eldan2023tinystories} to evaluate \methodsc attributions over the GPT-2 language model (\Cref{app:language_models}).
This dataset contains short stories synthetically generated by GPT-3.5 and GPT-4.
To compute component attributions for GPT-2, we specify a computation graph over $64,512$ components, which correspond to the set of all weight vectors, i.e., in every self-attention module and feed-forward module of the model.
See~\Cref{app:gpt2} for experiment details and findings.

\paragraph*{BoolQ.}
We use the BoolQ dataset \cite{clark2019boolq} to evaluate \methodsc attributions for the Phi-2 model \cite{li2023textbooks}.
Each example in this dataset comprises a passage of text, a question, and a binary answer. We evaluate the zero-shot performance of Phi-2 using the prompting and evaluation procedure from \citet{gao2023framework}\footnote{\url{https://github.com/EleutherAI/lm-evaluation-harness/}}.
Given the size of the Phi-2 model, we specify a computation graph over $55,552$ components, each corresponding to a contiguous block of $10$ weight vectors in every self-attention module and feed-forward module of the model.
See~\Cref{app:phi2} for experiment details and findings.

\subsection{Baselines}
\label{app:baselines}
In~\Cref{sec:eval}, we compare \methodsc against four baseline methods for estimating component attributions: Leave-One-Out (\texttt{LOO}), Gradient-times-parameters (\texttt{GP}), Neuron Conductance (\texttt{NC}), and Internal Influence (\texttt{II}).
Each baseline computes an attribution vector $\smash{\bm{w^{(z)}}} \in \mathbb{R}^{|C|}$ for a given example $z$ by assigning an ``importance'' score $\smash{w^{(z)}_j}$ to each component $c_j \in C$.
Then, as per~\Cref{eq:attribution_derived}, we estimate a component counterfactual $f_M(z, C')$ as the sum of importance scores of components in $C \setminus C'$, i.e., scores of components that are not ablated.
We describe each baseline in more detail below:
\begin{itemize}[leftmargin=*]
    \item \textbf{Leave-One-Out} (\texttt{LOO}): This method ablates each component $c_j \in C$ and sets the coefficient $\smash{\theta^{(z)}}_j$ to the change in model output $f_M(z)$ before and after ablation: $$\smash{w^{(z)}}_j = f_M(z, \{c_j\})-f_M(z, \emptyset)$$
    \item \textbf{Gradient-times-Parameters} (\texttt{GP}): This method approximates the leave-one-out estimate described above. Specifically, it estimates the leave-one-out effect of each component $c_j \in C$ using a first-order Taylor approximation of $f_M(z, \{c_j\})$ around $f_M(z, \emptyset)$: $$\smash{w^{(z)}_j =
    \nabla_{c_j} f_M(z, \emptyset) \cdot \delta_{c_j}}$$
    where $\delta_{c_j}$
    is the parameter-space change in $c_j$ induced by the ablation method of
    choice.
    \item \textbf{Neuron Conductance} (\texttt{NC}) \cite{dhamdhere2018important}: This method extends the Integrated Gradients method \cite{sundararajan2017axiomatic}---an input-space feature attribution method---to compute importance scores for each component $c_j \in C$. Intuitively, \texttt{NC} modifies the computation in Integrated Gradients in order to quantify the ``flow'' through each component $c_j \in C$. See Equation 3 in \cite{dhamdhere2018important} for a formal description.
    \item \textbf{Internal Influence} (\texttt{II}) \cite{leino2018influence}: Similar to \texttt{NC}, this method also adapts Integrated Gradients~\cite{sundararajan2017axiomatic} to compute importance scores. At a high level, \texttt{II} directly applies Integrated Gradients to layerwise activations by treating the output of each layer as an input to subsequent layers. See Definition 1 in \cite{leino2018influence} for a formal description.
\end{itemize}
We implement the first two baselines (\texttt{LOO} and \texttt{GP}) from scratch\footnote{Our code is available at \url{https://github.com/MadryLab/modelcomponents}} and use the \texttt{captum} library \cite{kokhlikyan2020captum} \footnote{Github repository: \url{https://github.com/pytorch/captum}} to implement \texttt{NC} and \texttt{II}.
As per \Cref{def:attribution}, we estimate the component counterfactual $f_M(z, C')$ using these baselines by setting the bias term $b^{(z)}$ to zero and taking the sum over attribution scores of components that are not ablated.

\subsection{Implementation details}
\label{app:implementation_details}

\paragraph{Sample size for component attribution estimation.}
The computational cost of our approach linearly scales with the sample size $m$ used to estimate component attributions (see~\Cref{alg:pseudo}).
Each sample in the component dataset $D^{(z)}$ corresponds to a single forward pass through the model $M$ in order to compute the counterfactual $f_M(z, C')$ \eqref{eq:ablation}, i.e., model output $f_M(z)$ after ablating a subset of components $C' \subset C$.
The setups $\{\text{A}, \text{B}, \text{C}\}$ considered in~\Cref{sec:eval} use sample size $m=\{50000, 100000, 200000\}$ respectively.
In~\Cref{app:sample_complexity}, we show that the sample size $m$ used in~\Cref{sec:eval} can be reduced by $2$-$5\times$, resulting in a direct speedup while only reducing the predictive power of \methodsc attributions by a small amount.

\paragraph{Data loading.}
We use the \texttt{FFCV} library\footnote{Github repository: \url{https://github.com/libffcv/ffcv}}~\citep{leclerc2022ffcv} to train and evaluate models. \texttt{FFCV} removes the data loading bottleneck for small models, gives a $3\text{-}4 \times$ improvement in throughput compared to standard PyTorch data loading.

\paragraph*{Speeding up regression.}
The second step of \methodsc---fitting component attributions to the component dataset \eqref{eq:dataset}---requires solving a linear regression problem (\Cref{eq:regression}) for each example $z$.
We parallelize this step by using the \texttt{fast-l1} package\footnote{Github repository: \url{https://github.com/MadryLab/fast_l1}}, a SAGA-based GPU solver for linear regression.

\paragraph{Computing resources.}
\label{app:resources}
We train our models and compute \methodsc attributions on a cluster of machines, each with $9$ NVIDIA A100 or V100 GPUs and $96$ CPU cores. We also use half-precision to increase training speed.

\section{Applying \methodsc to language models}
\label{app:language_models}

In~\Cref{sec:eval} and~\Cref{app:additional_evaluation}, we showed that our proposed method \methodsc attributions accurately estimate component counterfactuals \eqref{eq:ablation} on large-scale vision tasks across several datasets and model architectures.
In this section, we apply \methodsc to language models.
Specifically, we consider two experiments: (a) GPT-2~\citep{radford2019language} evaluated on the next-token prediction task and (b) Phi-2~\citep{li2023textbooks} evaluated on a zero-shot classification task.
In both cases, we show that \methodsc attributions accurately predict how model outputs change in response to component ablations.

\subsection{Evaluating GPT-2 on the TinyStories dataset}
\label{app:gpt2}

\paragraph{Task and model output function.}
We apply \methodsc to the next-token prediction task.
Following~\citet{park2023trak}, we interpret this task as a sequence as a $v$-way classification problem, where $v$ is the vocabulary size, and set the model output function to be the average correct-class margin \eqref{eq:margin} over all tokens in a given sequence.

\paragraph{Model and dataset.}
In this experiment, we consider the GPT-2 model\footnote{\url{https://huggingface.co/gpt2}}~\citep{radford2019language}, with a computation graph over $64,512$ components. These components correspond to the set of weight vectors in every self-attention module and feed-forward module in the model.
We evaluate model performance on the next-token prediction task using the TinyStories dataset\footnote{\url{https://huggingface.co/datasets/roneneldan/TinyStories}}~\citep{eldan2023tinystories}, where each sequence corresponds to a synthetically generated short story.

\paragraph{Computing \methodsc attributions.}
We apply \methodsc (without any modifications) to compute component attributions for a random subset of $1000$ examples in the TinyStories validation set using a component dataset of $200,000$ component counterfactuals \eqref{eq:dataset} and a ablation fraction of $\alpha = 2.5\%$.

\begin{figure}[!b]
    \centering
    \includegraphics[width=\textwidth]{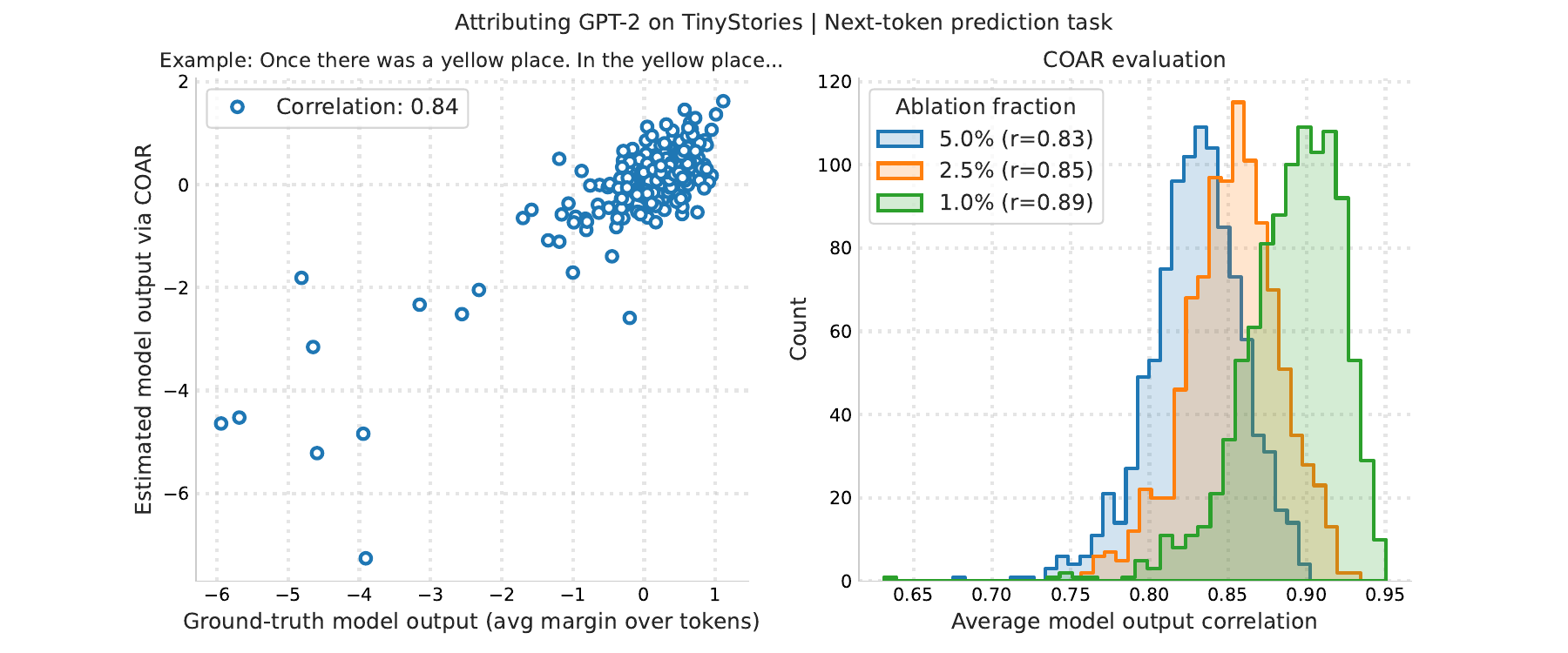}
    \caption{
        \textbf{Evaluating \methodsc on GPT-2.}
        We apply \methodsc to the GPT-2 model~\citep{radford2019language} on the TinyStories dataset~\citep{eldan2023tinystories}.
        The resulting component attributions are predictive of component counterfactuals.
        The left plot shows that component attributions can estimate the effect of ablating components on the average correct-class margin (over tokens in a sequence) of GPT-2 on a random TinyStories example with high correlation.
        The histograms in the right plot show that \methodsc attributions attain high average correlation for multiple values of ablation fraction $\alpha$.
    }
    \label{fig:gpt2}
\end{figure}

\paragraph{Evaluating \methodsc attributions.}
Similar to the results in~\Cref{sec:eval}, \methodsc attributions are predictive in the language modeling setting as well.
Specifically, these attributions accurately predict the effect of ablating components on the average correct-class margin of GPT-2 on examples from the TinyStories validation set.
In \Cref{fig:gpt2}a, we pick a random example $z$ from the TinyStories validation set and compute the correlation between ground-truth component counterfactuals $f_M(z, \cdot)$ and the corresponding estimate \eqref{eq:attribution_derived} using its \methodsc attributions $\bm{\theta^{(z)}}$, as defined in~\Cref{eq:correlation}.
In~\Cref{fig:gpt2}b, we plot a histogram over example-level correlations of $1000$ examples and find that \methodsc attributions attain an average correlation of $\{0.83, 0.85, 0.89\}$ with ground-truth component counterfactuals sampled using ablation fraction $\alpha=\{5\%, 2.5\%, 1\%\}$ respectively.

\subsection{Evaluating Phi-2 on the BoolQ dataset}
\label{app:phi2}

\paragraph{Task and model output function.}
We now turn to a reading comprehension task, where the goal is to answer a question given a passage of text.
We evaluate this classification task in a zero-shot manner: the language model is prompted with a passage of text and a question, and the goal is to output the correct answer from \{yes, no\}.
Like in vision tasks (\Cref{sec:eval}), we use the correct-class margin \eqref{eq:margin} as the model output function for this zero-shot binary classification task.

\paragraph{Model and dataset.}
We consider the Phi-2 model\footnote{\url{https://huggingface.co/microsoft/phi-2}}~\citep{li2023textbooks} and specify a computation graph over $55,552$ components. Here, each component corresponds to a contiguous block of $10$ weight vectors in the model.
We evaluate this model on the BoolQ dataset\footnote{\url{https://huggingface.co/datasets/google/boolq}}~\citep{clark2019boolq}, where each example consists of a passage of text, a question, and a binary \{yes, no\} answer. Using the prompting and evaluation procedure from the \citet{gao2023framework}\footnote{\url{https://github.com/EleutherAI/lm-evaluation-harness}}, Phi-2 attains an $83.6\%$ accuracy on this task.

\paragraph{Computing \methodsc attributions.}
Like in~\Cref{app:gpt2}, we apply compute \methodsc attributions for a random subset of $500$ examples in the BoolQ validation set using a component dataset of $m=100,000$ component counterfactuals \eqref{eq:dataset} and a ablation fraction of $\alpha = 0.025$.

\begin{figure}[!b]
    \centering
    \includegraphics[width=\textwidth]{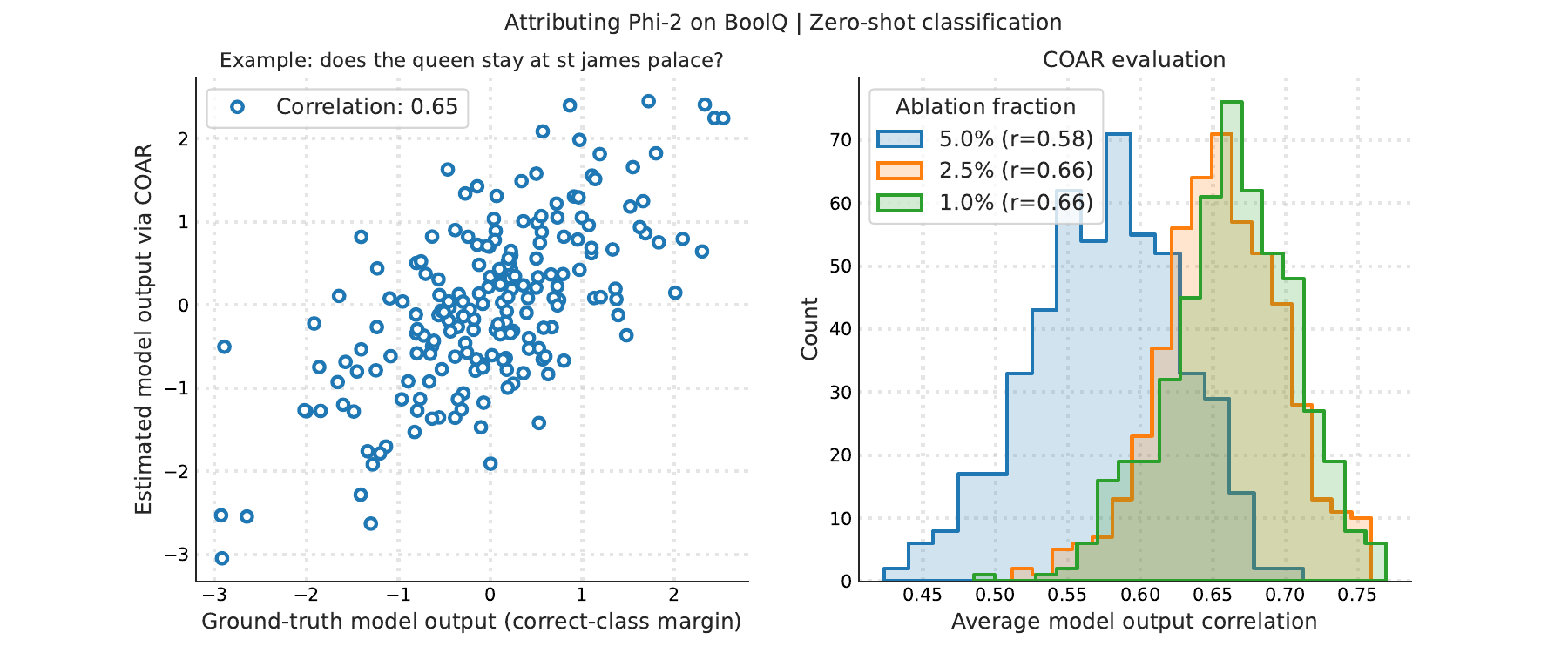}
    \caption{
        \textbf{Evaluating \methodsc on Phi-2.}
        We apply \methodsc to the Phi-2 model~\citep{javaheripi2023phi} on the BoolQ dataset~\citep{clark2019boolq}.
        The resulting component attributions are predictive of component counterfactuals.
        The left plot shows that component attributions can estimate the effect of ablating components on the average correct-class margin of Phi-2 on a random BoolQ example with high correlation.
        The histograms in the right plot show that \methodsc attributions attain high average correlation for multiple values of ablation fraction $\alpha$.
    }
    \label{fig:phi2}
\end{figure}

\paragraph{Evaluating \methodsc attributions.}
We find that \methodsc attributions are predictive of unseen component counterfactuals on this task as well.
\Cref{fig:phi2}a plots the correlation between ground-truth component counterfactuals $f_M(z, \cdot)$ and the corresponding \methodsc estimate \eqref{eq:attribution_derived} of a random BoolQ example $z$.
The histograms in~\Cref{fig:phi2}b show that \methodsc attributions attain an average correlation of $\{0.58, 0.66, 0.66\}$ with ground-truth component counterfactuals sampled using ablation fraction $\alpha=\{5\%, 2.5\%, 1\%\}$ respectively.

\clearpage

\section{Additional evaluation of \methodsc}
\label{app:additional_evaluation}
In this section, we first show that \methodsc learns accurate component attributions on additional datasets, model architectures, and tasks (\Cref{app:dataset_eval,app:arch_eval,app:setting_eval}).
This supplements our findings in~\Cref{sec:eval}, where we showed that \methodsc learns component attributions that accurately predict component counterfactuals \eqref{eq:ablation} on three image classification setups: CIFAR-10 ResNet-18, ImageNet ResNet-50, and ImageNet ViT-B/16.
Then, we show that \methodsc attributions retain its predictive power when estimated with fewer samples (\Cref{app:sample_complexity}) or with different ablation fractions (\Cref{app:alpha_eval}).
Finally, we supplement our example-level evaluation of \methodsc attributions in~\Cref{sec:eval} with additional example-level comparisons of ground-truth component counterfactuals and attribution-based estimates (\Cref{app:baseline_eval}).

\subsection{Evaluating \methodsc on additional datasets}
\label{app:dataset_eval}
\begin{figure}[b]
    \centering
    \includegraphics[width=\textwidth]{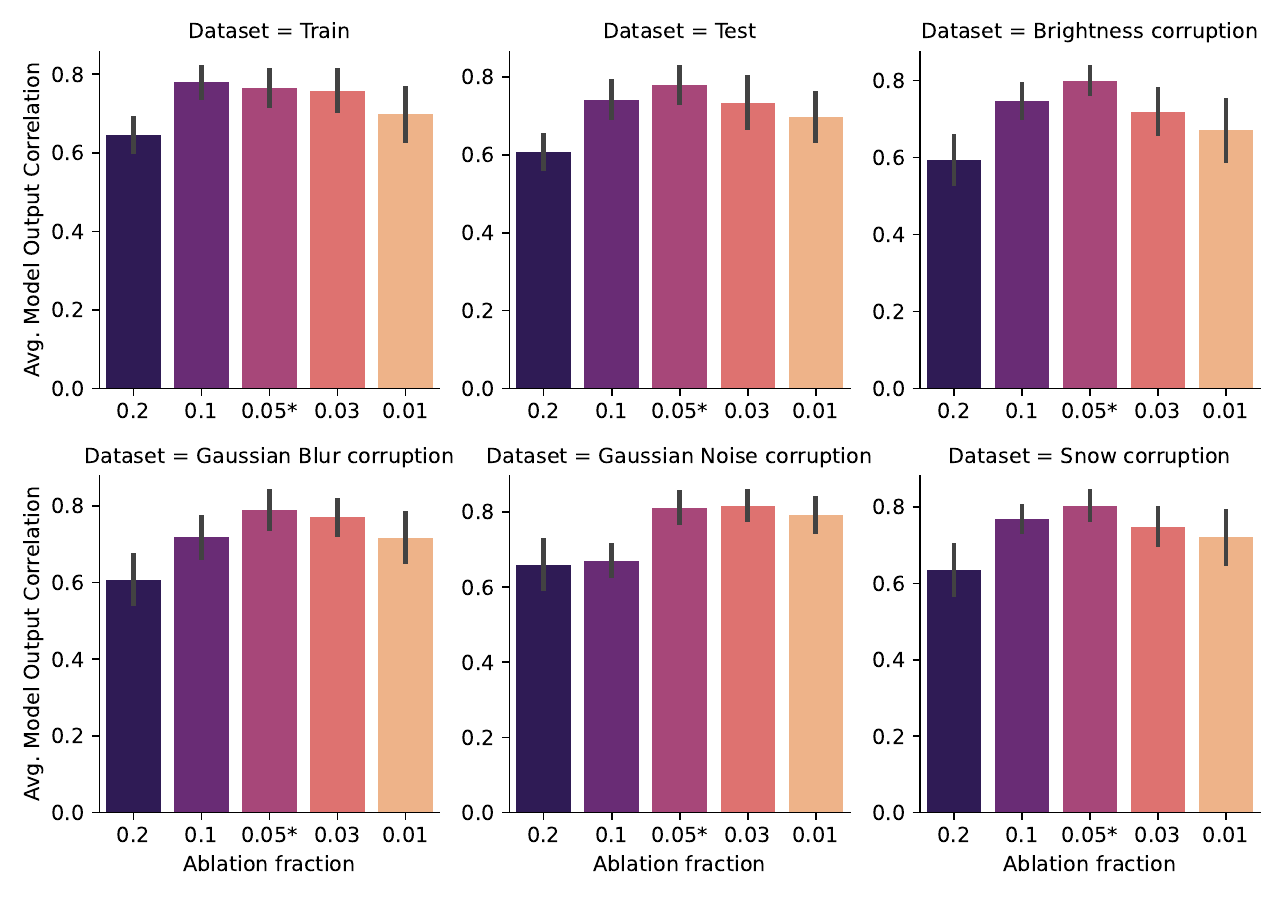}
    \caption{
        \textbf{Do \methodsc attributions generalize to out-of-distribution examples?}
        \methodsc attributions remain predictive on the CIFAR-10 training set and on six corrupted versions of the CIFAR-10 test set~\cite{hendrycks2019benchmarking} over a range of ablation fractions $\alpha$. See~\Cref{app:dataset_eval} for more details.
    }
    \label{fig:ood_data}
\end{figure}

Our experiments in \Cref{sec:eval} evaluated the predictiveness of \methodsc attributions corresponding to in-distribution test examples from the CIFAR-10 and ImageNet datasets.
Now, we show that \methodsc attributions remain predictive on training examples as well as out-of-distribution examples.
Specifically, we apply \methodsc to compute attributions of ResNet-18 predictions on the CIFAR-10 training set and on six corrupted versions of the CIFAR-10 test set~\cite{hendrycks2019benchmarking}.
as shown in~\Cref{fig:ood_data}, \methodsc attributions exhibit high correlation on average (between $0.6$ and $0.8$) depending on the ablation fraction $\alpha$ used to ablate random $\alpha$-fraction sized components subsets.
Note that the correlation is maximum when $\alpha=0.05$ because the component attributins are estimated with the same ablation fraction, i.e., $\alpha_\text{train} = 0.05$.

\subsection{Evaluating \methodsc on additional model architectures}
\label{app:arch_eval}
\begin{figure}[b]
    \centering
    \includegraphics[width=\textwidth]{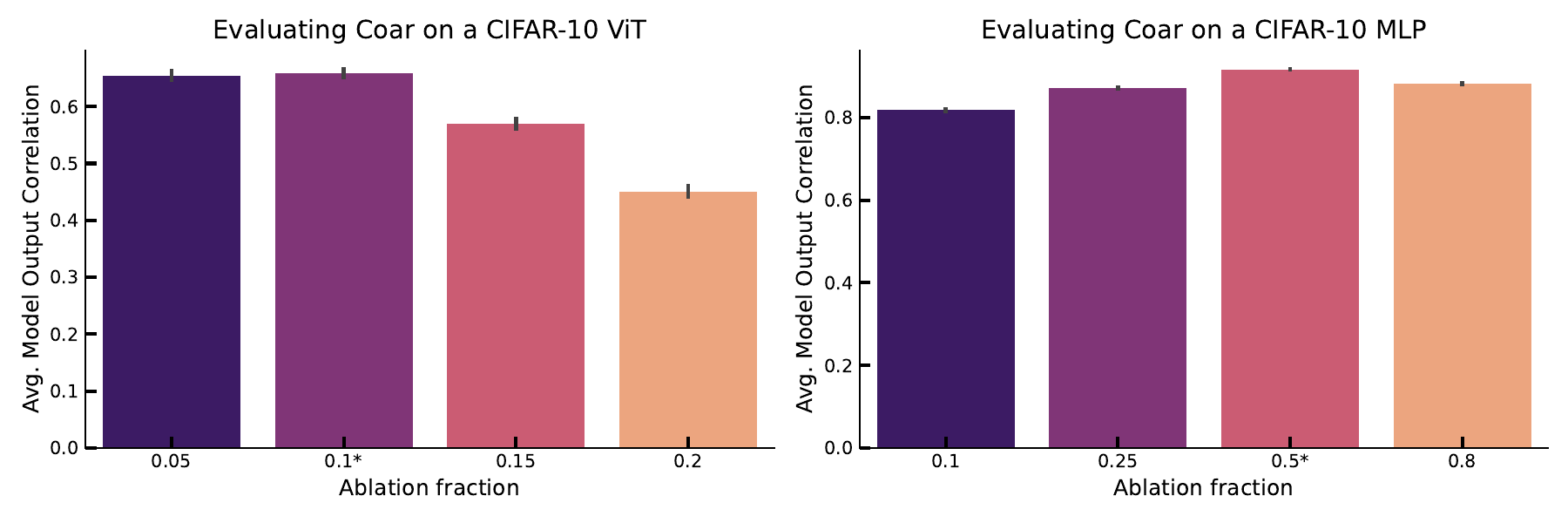}
    \caption{
        \textbf{Do \methodsc attributions generalize to other model architectures?}
        \methodsc attributions yield accurate estimates of component counterfactuals on two additional model architectures: a ViT-based model (left) and a one-layer fully-connected network (right) trained on CIFAR-10. See~\Cref{app:arch_eval} for more details.
    }
    \label{fig:arch}
\end{figure}

Recall that \methodsc is model-agnostic in that it is not tied to any specific model architecture.
In~\Cref{sec:eval}, we applied \methodsc to ResNets trained on CIFAR-10 and ImageNet and a ViT-B/16 model trained on ImageNet.
In this section, we apply \methodsc to two additional model architectures: a ViT model trained on CIFAR-10 ($83\%$ accuracy) and a one-layer fully-connected network trained on CIFAR-10 ($56\%$ accuracy).
\Cref{fig:arch} shows that \methodsc attributions on both architectures yield accurate estimates of how model outputs change in response to ablating random $\alpha$-fraction sized components subsets, with correlation $0.65$ and $0.85$ for the ViT and MLP models when $\alpha=\alpha_\text{train}$ respectively.

\subsection{Evaluating \methodsc on additional tasks}
\label{app:setting_eval}
\begin{figure}[!h]
    \centering
    \includegraphics[width=\textwidth]{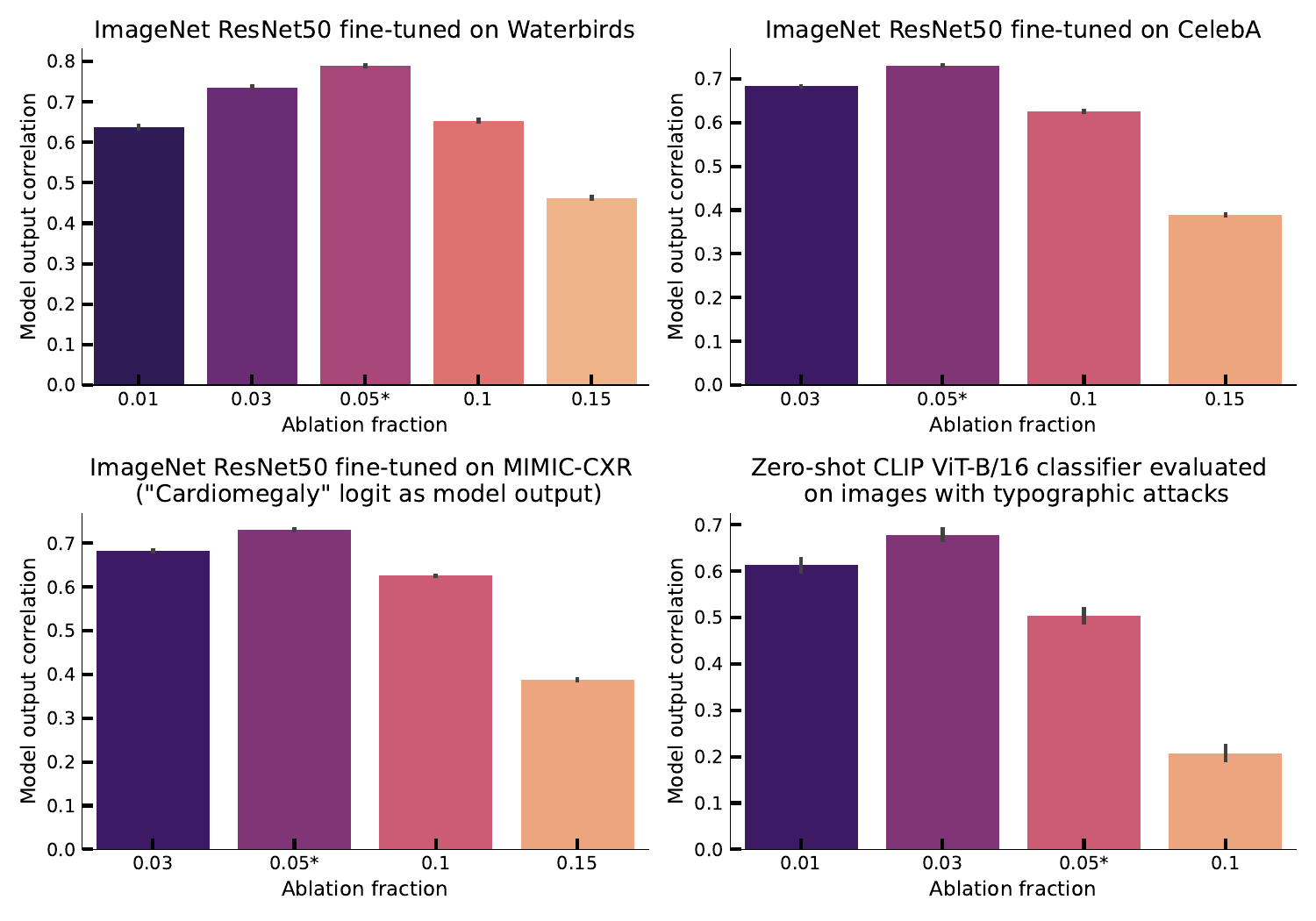}
    \caption{
        \textbf{Evaluating \methodsc attributions on additional tasks.}
        We find that component attributions estimated using \methodsc are predictive on four additional tasks: fine-tuning ImageNet ResNet50 on Waterbirds, CelebA and MIMIC-CXR, and a zero-shot CLIP ViT-B/16 classification task on a dataset containing typographic attacks (\Cref{subsec:typographic}).
        Note that the MIMIC-CXR setting uses the logit of the ``Cardiomegaly'' class as the model output function.
        See~\Cref{app:setting_eval} for additional information about these tasks.
    }
    \label{fig:tasks}
\end{figure}

We now evaluate \methodsc attributions on four additional tasks:
\begin{itemize}
\item First, we apply \methodsc to pre-trained ImageNet ResNet50 model fine-tuned on two datasets---Waterbirds and CelebA---that we use in~\Cref{subsec:subpops}---see first row of~\Cref{fig:tasks}.
We find that \methodsc attributions are predictive on both datasets, attaining higher correlation with ground-truth component counterfactuals when $\alpha$ is closer to $\alpha_\text{train}=0.05$.
\item Second, we apply \methodsc to a pre-trained ImageNet ResNet50 model fine-tuned on MIMIC-CXR~\cite{johnson2019mimic}, a dataset of labeled chest radiographs. In this case, we set the model output function to be the logit of the ``Cardiomegaly'' class instead of correct-class margin that we use in~\Cref{sec:eval}. \Cref{fig:tasks} shows that \methodsc attributions attain a correlation of $0.7$ and $0.6$ with ground-truth logits when $\alpha=\alpha_\text{train}=0.05$ and $\alpha=0.10$ respectively.
\item The fourth plot in~\Cref{fig:tasks} corresponds to the CLIP setting considered in~\Cref{sec:applications}. In this setting, we take the zero-shot CLIP ViT-B/16 classifier and evaluate it on a dataset of images with and without typographic attacks~\cite{materzynska2022disentangling}. As shown in the plot, the correlation between \methodsc attributions and ground-truth margins is close to $0.7$ when $\alpha=\alpha_\text{train}=0.03$, i.e., ablating $3\%$ of the components in the CLIP model.
\end{itemize}

\subsection{Comparing \methodsc attributions estimated with different ablation fractions}
\label{app:alpha_eval}
\begin{figure}[b]
    \centering
    \includegraphics[width=\textwidth]{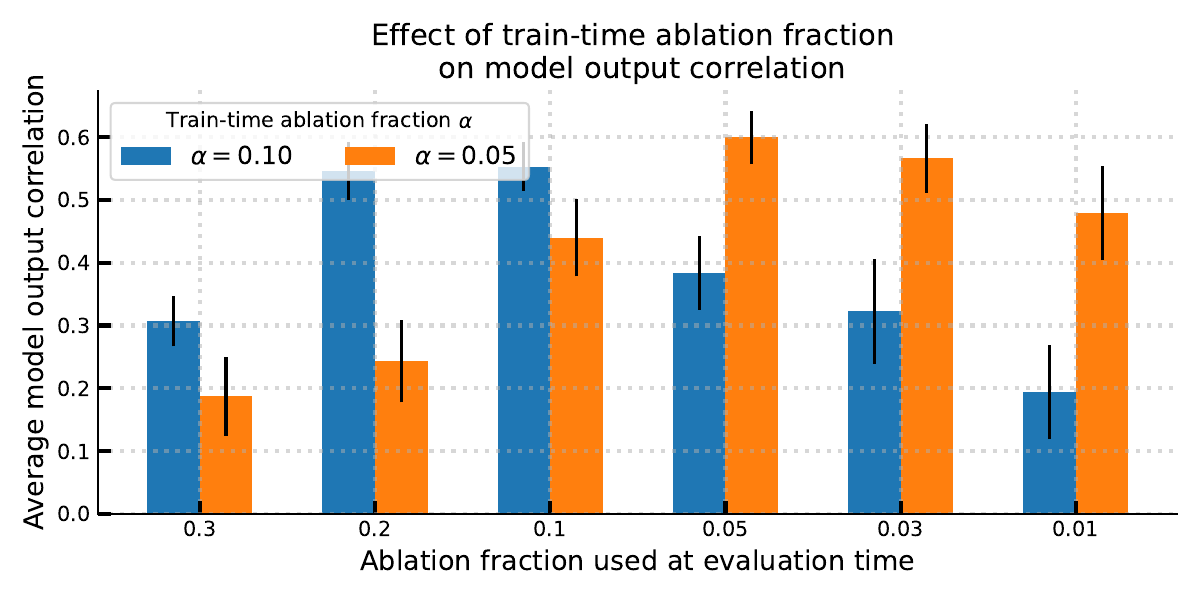}
    \caption{
        \textbf{Comparing \methodsc attributions estimated with different ablation fractions $\alpha$.}
        \methodsc attributions estimated with different ablation fractions $\alpha_\text{train}$ attain a different correlation ``profile'' over $\alpha$ at test time.
        The correlation between ground-truth component counterfactuals and attribution-based estimates  is higher for attributions estimated with $\alpha_\text{train}=0.05$ when $\alpha$ is small, and higher for attributions estimated with $\alpha_\text{train}=0.10$ when $\alpha$ is large.
        This empirically shows that \methodsc attributions are more predictive on component counterfactuals that are ``similar'' to the ones used to fit them---i.e., when $\alpha_\text{test} \approx \alpha_\text{train}$.
        See~\Cref{app:alpha_eval} for more details.
    }
    \label{fig:vary_alpha}
\end{figure}

We now analyze how changing the ablation fraction $\alpha_\text{train}$ used to fit \methodsc attributions affects their predictiveness over different ablation fractions at test time.
Specifically, we consider the ImageNet ResNet-50 setting from~\Cref{sec:eval} and compute two sets of \methodsc attributions, corresponding to two values of $\alpha_\text{train}$: $0.05$ and $0.10$.
Then, for each of these two sets of attributions, we evaluate its correlation with ground-truth component counterfactuals over a range of ablation fractions $\alpha$.
As shown in~\Cref{fig:vary_alpha}, the correlation ``profile'' over $\alpha$ depends on the value of $\alpha_\text{train}$ used to fit the attributions.
When $\alpha$ is small, the correlation is higher for attributions estimated with $\alpha_\text{train}=0.05$.
Analogously, when $\alpha$ is large, the correlation is higher for attributions estimated with $\alpha_\text{train}=0.10$.
This is because the component attributions fare better as counterfactual predictors on component counterfactuals that are ``similar'' to the ones used to fit them---i.e., when $\alpha_\text{test} \approx \alpha_\text{train}$.

\subsection{Comparing \methodsc attributions estimated with different sample sizes}
\label{app:sample_complexity}
\begin{figure}[!b]
    \centering
    \includegraphics[width=\textwidth]{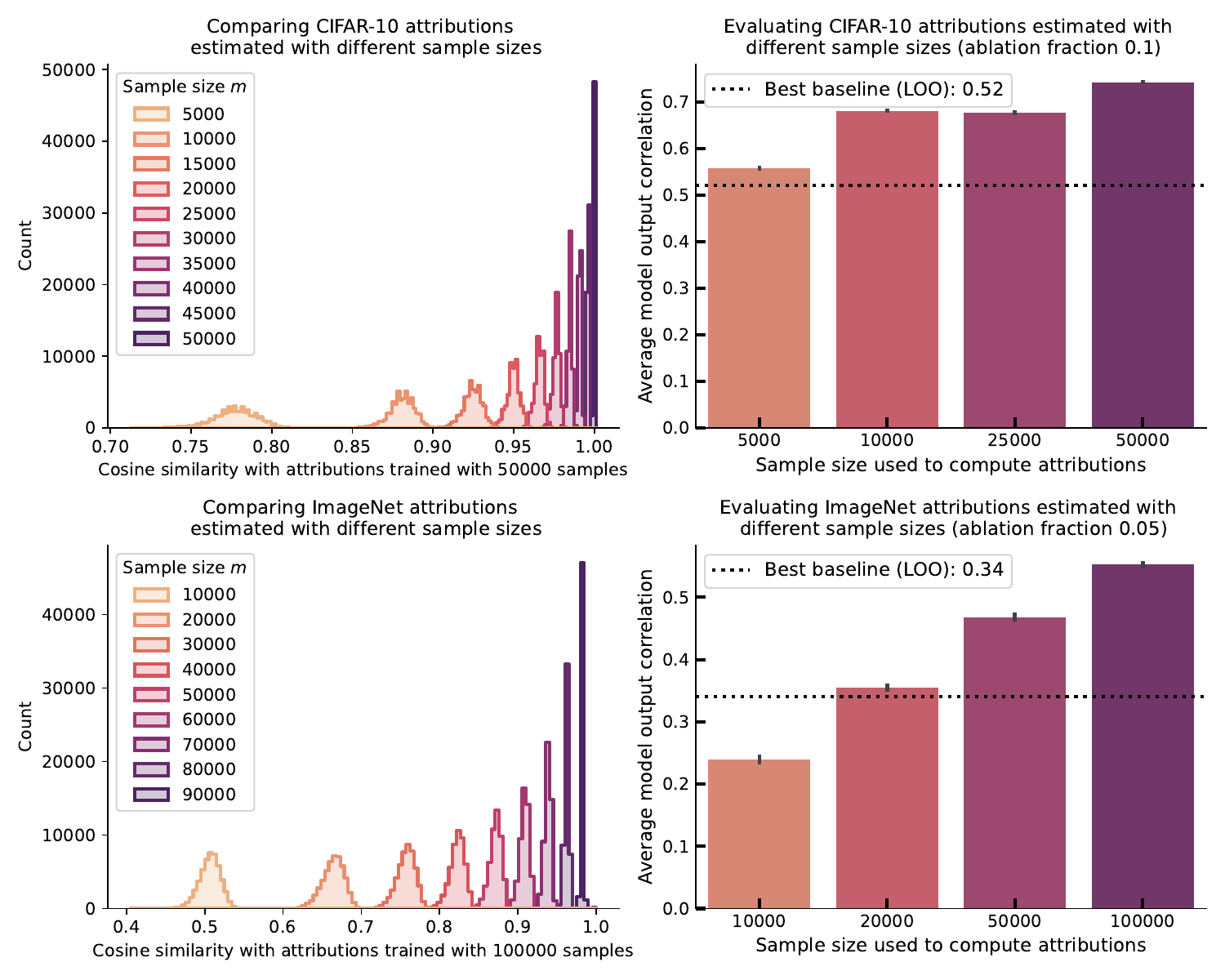}
    \caption{
        \textbf{Comparing \methodsc attributions estimated with different sample sizes.}
        \methodsc attributions for CIFAR-10 ResNet-18 and ImageNet ResNet-50 (Setup A and B respectively in~\Cref{sec:eval})
        estimated with smaller sample sizes $m$ are still predictive of component counterfactuals.
        On the left, we show that \methodsc attributions estimated with sample size $m$ and $m/k$ have high cosine similarity on average, with the similarity increasing as $k$ decreases.
        On the right, we show that decreasing the sample size $m$ by a factor of $k \in \{2, 5, 10\}$ does not significantly affect the correlation between \methodsc attributions and ground-truth component counterfactuals.
        In particular, \methodsc outperforms the best-performing baseline (\texttt{LOO}) even with $10\times$ fewer samples on CIFAR-10 (top row) and $5\times$ fewer samples on ImageNet (bottom row).
    }
    \label{fig:samplesize}
\end{figure}

In~\Cref{sec:eval}, we computed \methodsc attributions using sample sizes $m=50000$ for the ResNet-18 model trained on CIFAR-10 and $m=100000$ for the ResNet-50 model trained on ImageNet.
Recall that the sample size $m$ here corresponds to the number of component counterfactuals used to fit the component attributions.
In this section, we vary the sample size $m$ and show that \methodsc attributions remain predictive even when trained on $k \times$ fewer examples, where $k \in \{2, 5, 10\}$.
Specifically, the left column of~\Cref{fig:samplesize} shows that \methodsc attributions estimated on CIFAR-10 and ImageNet data with sample size $m$ and $m/k$ have high cosine similarity on average, with the similarity increasing as $k$ decreases.
The right column of~\Cref{fig:samplesize} shows that decreasing the sample size $m$ by a factor of $k \in \{2, 5, 10\}$ does not significantly impact the correlation between \methodsc attributions and ground-truth component counterfactuals.
For example, reducing the sample size by $5\times$ only reduces the correlation from $0.7$ to $0.65$ in the CIFAR-10 ResNet-18 setting.
Additionally, we observe that \methodsc attributions fare better than attributions estimated with the best-performing baseline (\texttt{LOO}) even when trained on $10\times$ fewer examples on CIFAR-10 and $5\times$ fewer examples on ImageNet.

\subsection{Analyzing \methodsc attributions at the example level}
\label{app:baseline_eval}
\begin{figure}[!b]
    \centering
    \includegraphics[width=\textwidth]{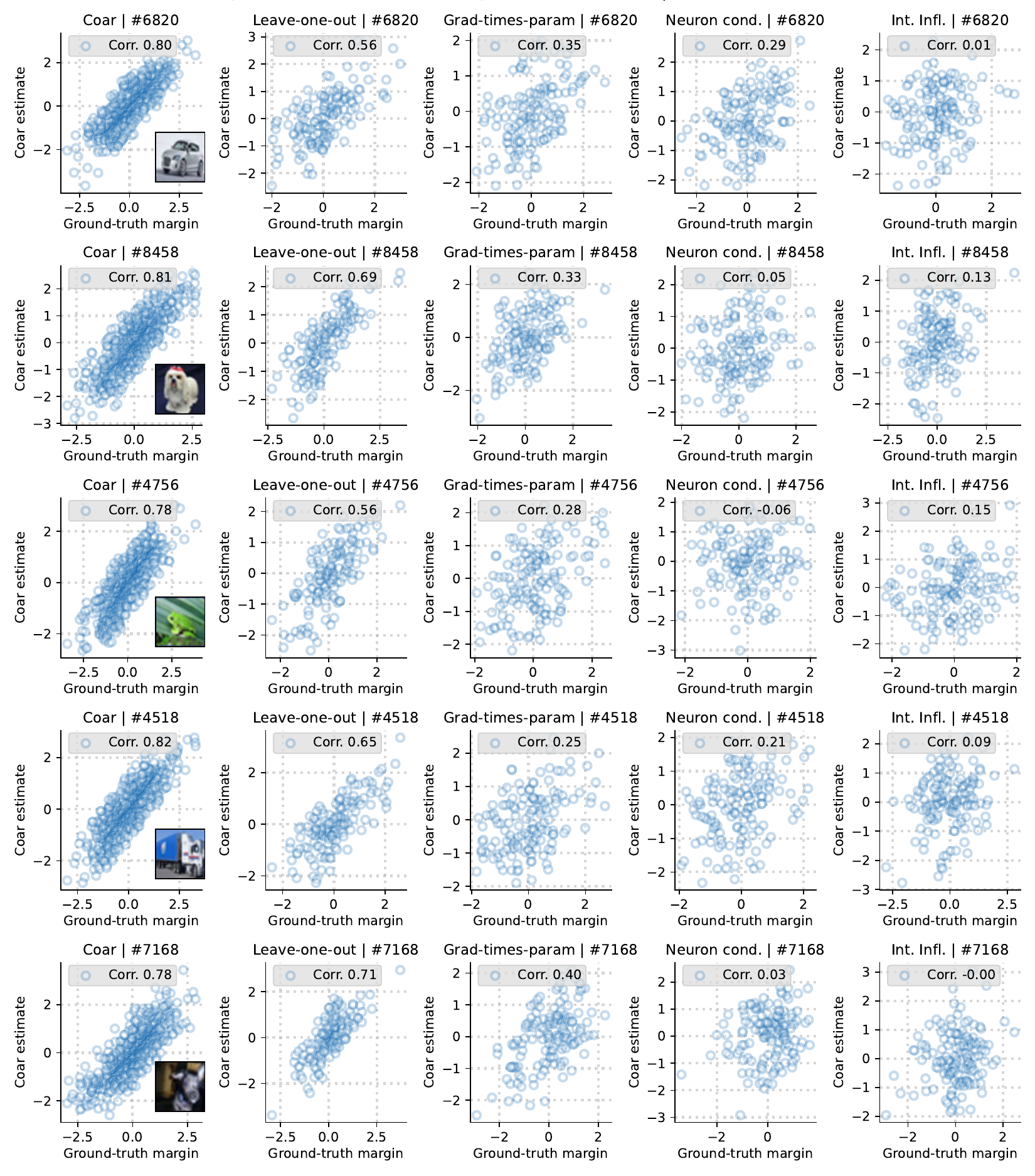}
    \caption{
        \textbf{Additional example-level evaluation of component attributions on CIFAR-10.}
        Each row corresponds to a different example $z$ randomly picked from the CIFAR-10 test set and each column corresponds to a different attribution method.
        The left-most subfigure in each row shows that \methodsc attributions and the corresponding ground-truth component counterfactuals exhibit high correlation on example $z$.
        In comparison, the other subfigures in each row, one for baseline method, consistently exhibit lower correlation.
        See \Cref{app:sample_complexity} for more details.
    }
    \label{fig:example_eval_cifar}
\end{figure}

\begin{figure}[!b]
    \centering
    \includegraphics[width=\textwidth]{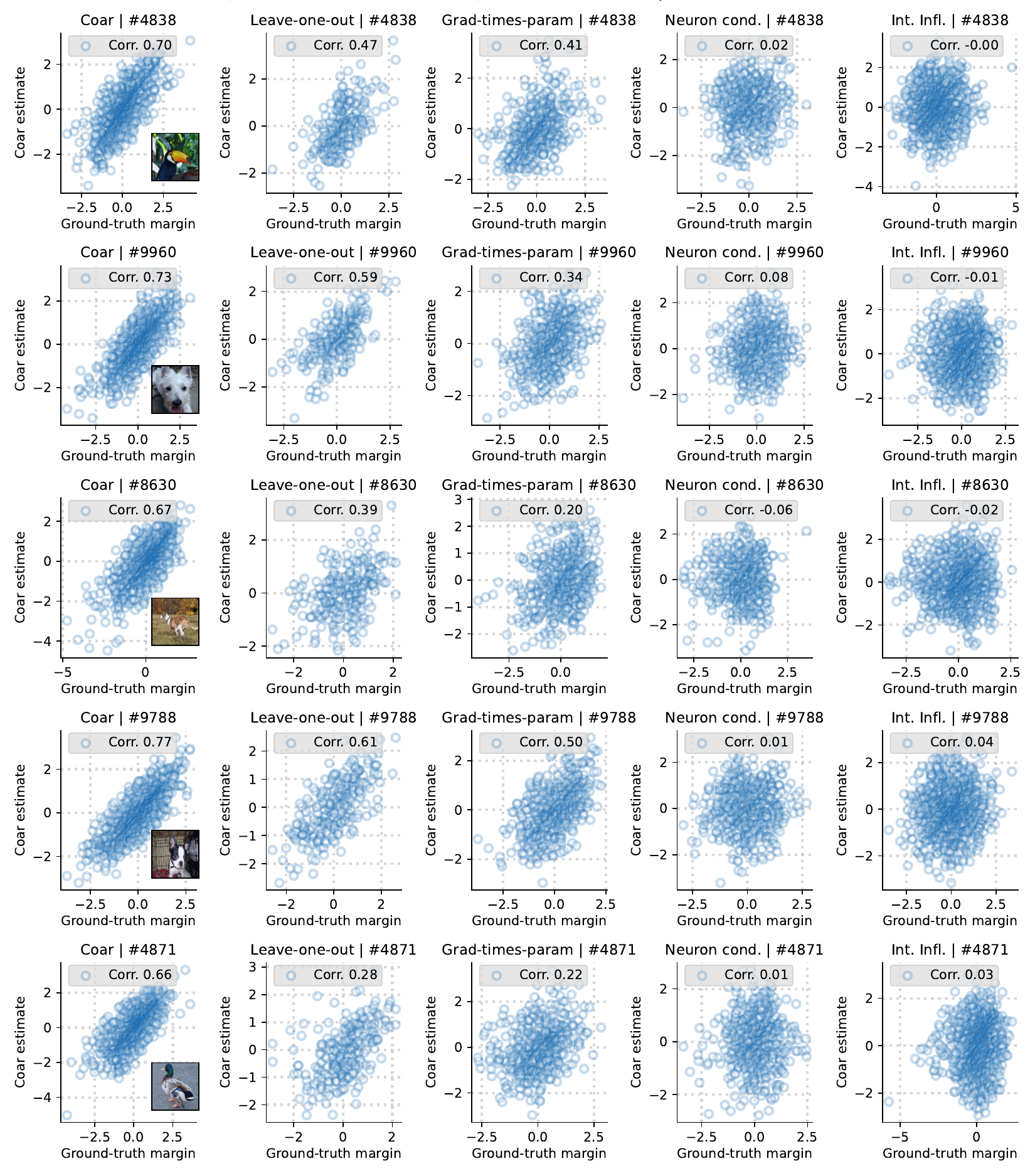}
    \caption{
        \textbf{Additional example-level evaluation of component attributions on ImageNet.}
        Similar to the results in~\Cref{fig:example_eval_cifar}, each row corresponds to a different example $z$ randomly picked from the ImageNet test set.
        The left-most subfigure in each row shows that \methodsc attributions and the corresponding ground-truth component counterfactuals exhibit high correlation on example $z$.
        In comparison, the other subfigures in each row, corresponding to a baseline method, consistently exhibit worse correlation.
        See \Cref{app:sample_complexity} for more details.
    }
    \label{fig:example_eval_in}
\end{figure}

To supplement our evaluation in~\Cref{sec:eval}, we provide additional example-level scatterplot comparisons between ground-truth component counterfactuals and the corresponding estimates obtained using component attributions estimated with \methodsc and all baselines from~\Cref{sec:eval}.
We plot these comparisons on CIFAR-10 examples in~\Cref{fig:example_eval_cifar} and on ImageNet examples in~\Cref{fig:example_eval_in}.
Our findings further substantiate that \methodsc attributions exhibit higher correlation with ground-truth component counterfactuals than all four baseliens on both CIFAR-10 and ImageNet.

\subsection{Qualitatively analyzing \methodsc attributions}
\label{app:qualitative}
\begin{figure}[!b]
    \centering
    \includegraphics[width=\textwidth]{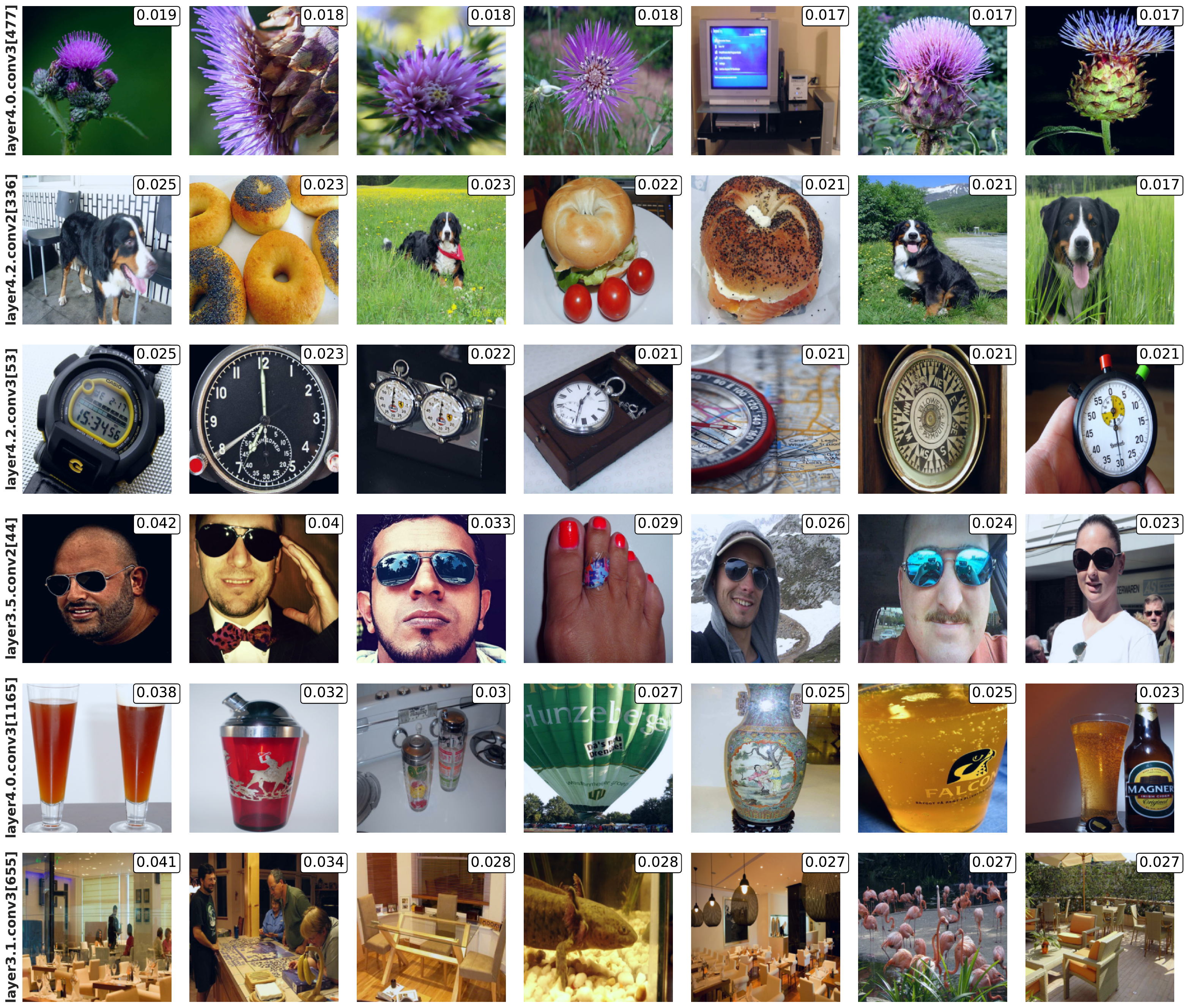}
    \caption{
        \textbf{Visualizing component-specific attributions across examples.}
        We sample a random set of components from the ImageNet ResNet-50 model (setup B in~\Cref{sec:eval}) and visualize the examples with the most positive attributions for each component.
        In general, the examples with the most positive attributions for a given component exhibit visual similarity at different levels of granularity.
        For example, the first, third and fifth row in~\Cref{fig:vis1} show that the examples with the most positive attributions for \texttt{layer4.0.conv3[477]} and \texttt{layer4.2.conv3[53]} contain purple flowers, watch faces, and glass-shaped objects respectively.
        However, consistent with recent work on superposition in deep networks~\cite{elhage2022toy}, we observe that some components such as \texttt{layer4.2.conv2[336]} (second row) and \texttt{layer3.1.conv3[655]} (last row) can surface dissimilar subsets of examples or do not readily map to a single semantic concept.
    }
    \label{fig:vis1}
\end{figure}

We qualitatively analyze \methodsc attributions using two visualization techniques:

\paragraph{Visualizing component-specific attributions across examples.}
Given examples $\{z_1, \ldots, z_n\}$ with corresponding component attributions $\{\theta^{(z_1)}, \ldots, \theta^{(z_n)}\}$, we analyze how the attribution estimates of individual components vary across the set of examples.
Specifically, for a component $c_i \in C$, we visualize the examples with the most positive attribution values $\smash{\theta^{(z)}_i}$ for component $c_i$.
In this experiment, we visualize a random subset of components from the ImageNet ResNet-50 model (setup B in~\Cref{sec:eval}).
As shown in~\Cref{fig:vis1}, the examples with the most positive attributions for a given component exhibit high visual similarity at different levels of granularity:
\begin{itemize}
    \item The first, third and fifth row in~\Cref{fig:vis1} show that the examples with the most positive attributions for \texttt{layer4.0.conv3[477]} and \texttt{layer4.2.conv3[53]} contain purple flowers, watch faces, and glass-shaped objects respectively.
    \item However, consistent with recent work on superposition in deep networks~\cite{elhage2022toy}, we observe that some components such as \texttt{layer4.2.conv2[336]} in the second row as well as \texttt{layer3.1.conv3[655]} in the last row can surface dissimilar subsets of examples and do not readily map to a single semantic concept.
\end{itemize}

\begin{figure}[!b]
    \centering
    \includegraphics[width=\textwidth]{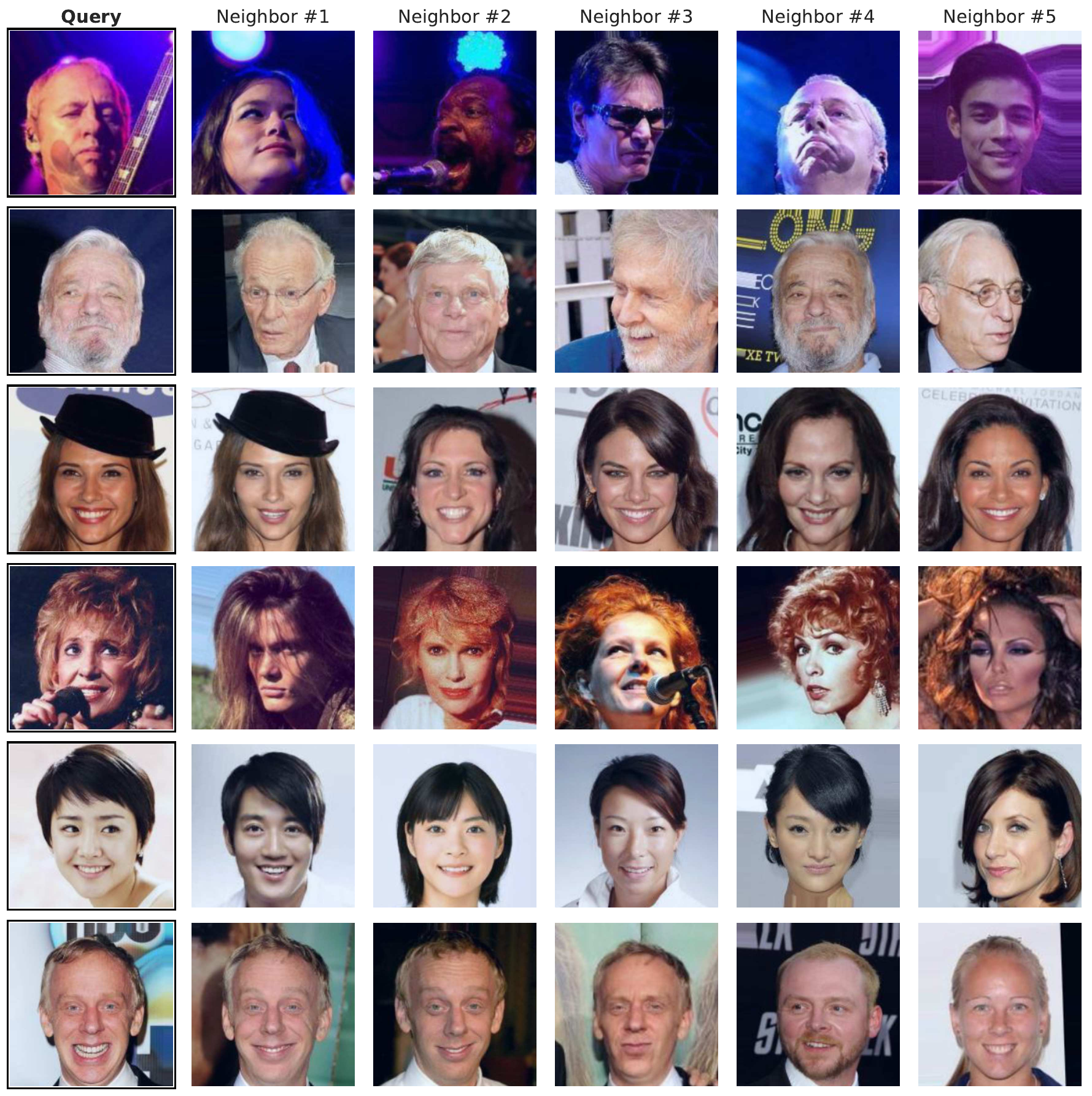}
    \caption{
        \textbf{Visualizing nearest neighbors in \methodsc attribution space.}
        We also use component attributions as feature embeddings in order to visualize the five nearest neighbors of examples from the CelebA dataset in ``component attribution'' space.
        Intuitively, this technique allows us to identify examples on which model outputs change similarly in response to component ablations.
        In general, we observe that the nearest neighbors of a given example in attribution space high visual similarity, e.g, similar facial attributes such as background (first row), hair color (second and fourth row), accessories (third row), or even the same person in different poses (last row).
    }
    \label{fig:vis2}
\end{figure}

\paragraph{Visualizing nearest neighbors in attribution space.} We also use component attributions as feature embeddings in order to visualize the nearest neighbors of a given example in ``component attribution'' space.
Intuitively, this technique allows us to identify examples on which model outputs change similarly in response to component ablations.
In this experiment, we visualize a random subset of examples from the CelebA dataset along with their $5$ nearest neighbors using \methodsc attributions of a fine-tuned ImageNet ResNet-50 model.
\Cref{fig:vis2} shows that the nearest neighbors of a given example in attribution space high visual similarity, i.e., similar facial attributes such as background (first row), hair color (second and fourth row), accessories (third row), or even the same person in different poses (last row).

\section{Additional evaluation of \editsc}
\label{app:additional_editing}
We use \editsc in five different editing tasks: correcting misclassifications (\S\ref{subsec:examples}); forgetting a class (\S\ref{subsec:class_forget}); improving subpopulation robustness (\S\ref{subsec:subpops}); localizing backdoor attacks (\S\ref{subsec:backdoor}); and improving robustness to typographic attacks (\S\ref{subsec:typographic}).
In this section, we provide additional details and/or supplementary experiments for each task.

\subsection{Editing individual predictions}
\label{app:example_level}

\paragraph{Experiment details.}
In~\Cref{subsec:examples}, we use \editsc to correct misclassifications of a ResNet-50 model on ImageNet examples.
In this experiment, we set the ``target'' example to be a misclassified ImageNet example and the ``reference'' example to a set of $50$ randomly selected ImageNet examples.
Then, we use these examples to identify and ablate components \eqref{eq:c-edit} that increase the correct-class margin \eqref{eq:margin} of the target example without impacting the average margin over the reference examples.

\begin{figure}[!b]
    \centering
    \includegraphics[width=0.8\textwidth]{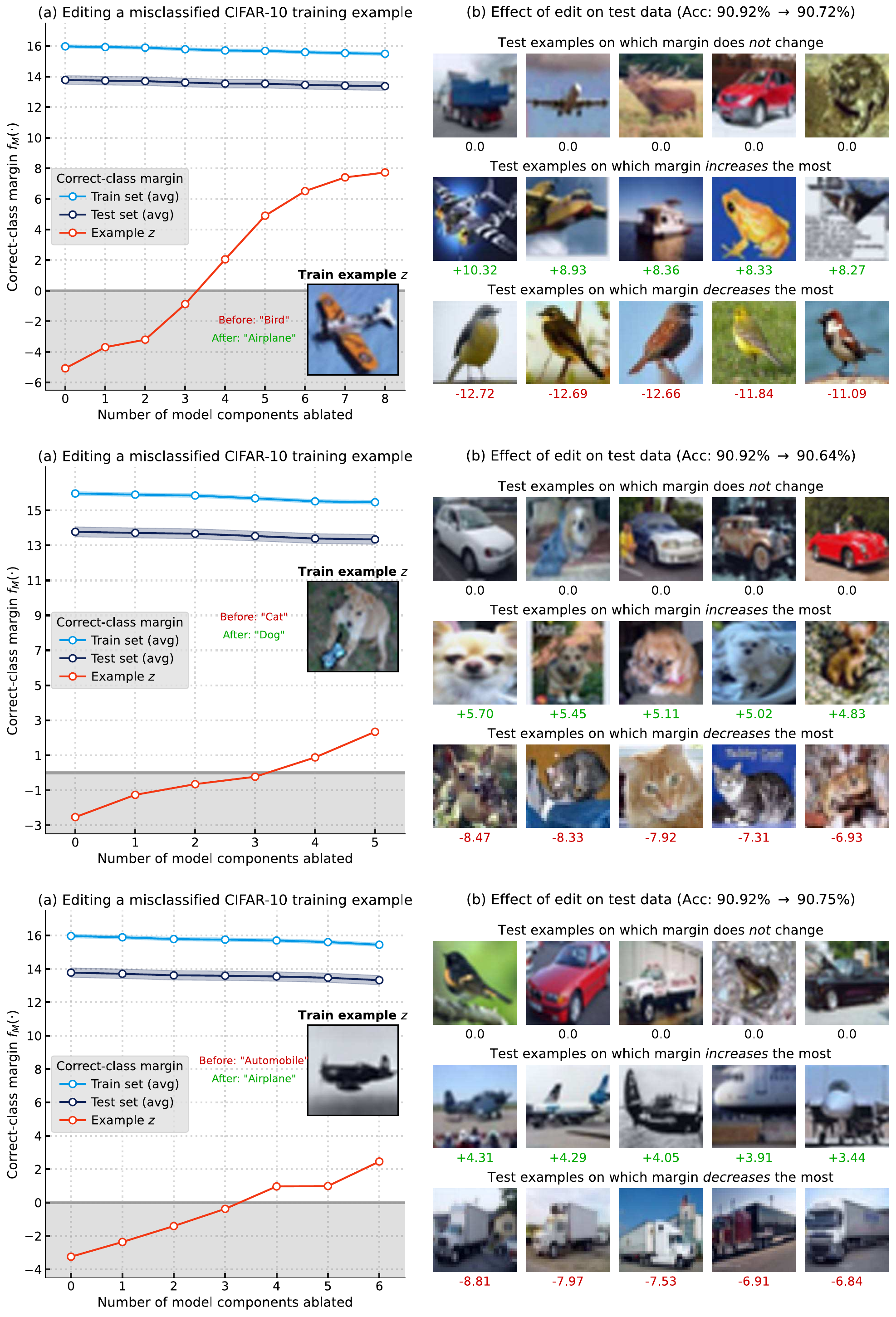}
    \caption{
        \textbf{Correcting misclassified CIFAR-10 examples via \editsc.}
        We reproduce the \editsc experiment from~\Cref{subsec:examples} on the CIFAR-10 dataset.
        Specifically, each row corresponds to CIFAR-10 test example that is misclassified by a ResNet-18 model.
        The left subplot in each row shows how applying \editsc (by ablating components \eqref{eq:c-edit}) increases the correct-class margin \eqref{eq:margin} of the misclassified example without impacting the average margin over the train or test set.
        The right subplot reports the drop in overall test accuracy and visualizes examples with correct-class margins that change the most or least due to the edit.
    }
    \label{fig:editing_cifar10_bottomk}
\end{figure}

\begin{figure}[!b]
    \centering
    \includegraphics[width=0.75\textwidth]{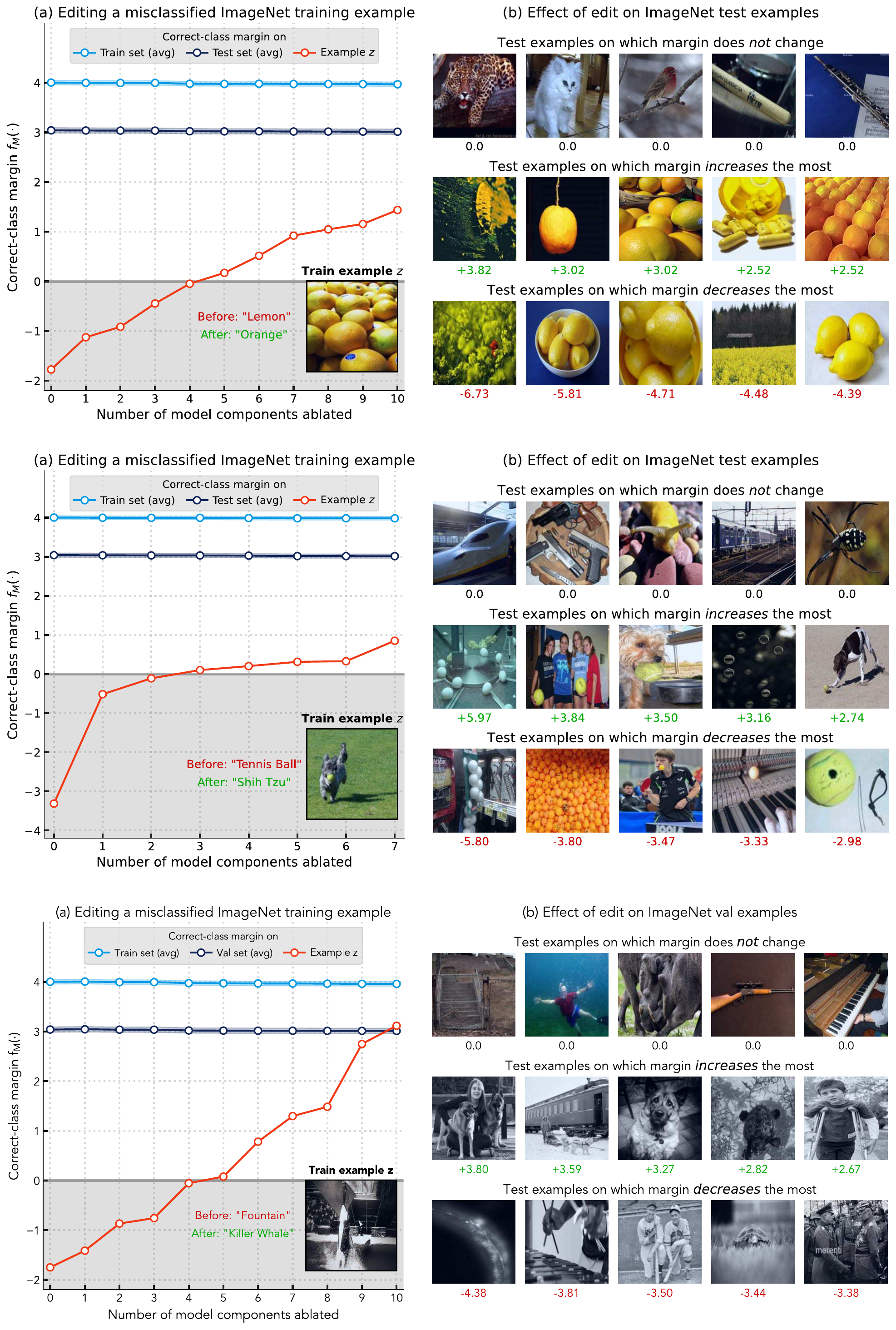}
    \caption{
        \textbf{Correcting misclassified ImageNet examples via \editsc.}
        We reproduce the \editsc experiment from~\Cref{subsec:examples} on additional ImageNet examples (one per row) misclassified by a ResNet-50 model.
        The left subplot shows that applying \editsc (by ablating components \eqref{eq:c-edit}) increases the correct-class margin \eqref{eq:margin} of the misclassified example without impacting the average margin over the train or test set.
        (Right)
        The right subplot visualizes examples with margins that change the most or least due to the edit.
    }
    \label{fig:editing_in_bottomk}
\end{figure}

\begin{figure}[!b]
    \centering
    \includegraphics[width=0.85\textwidth]{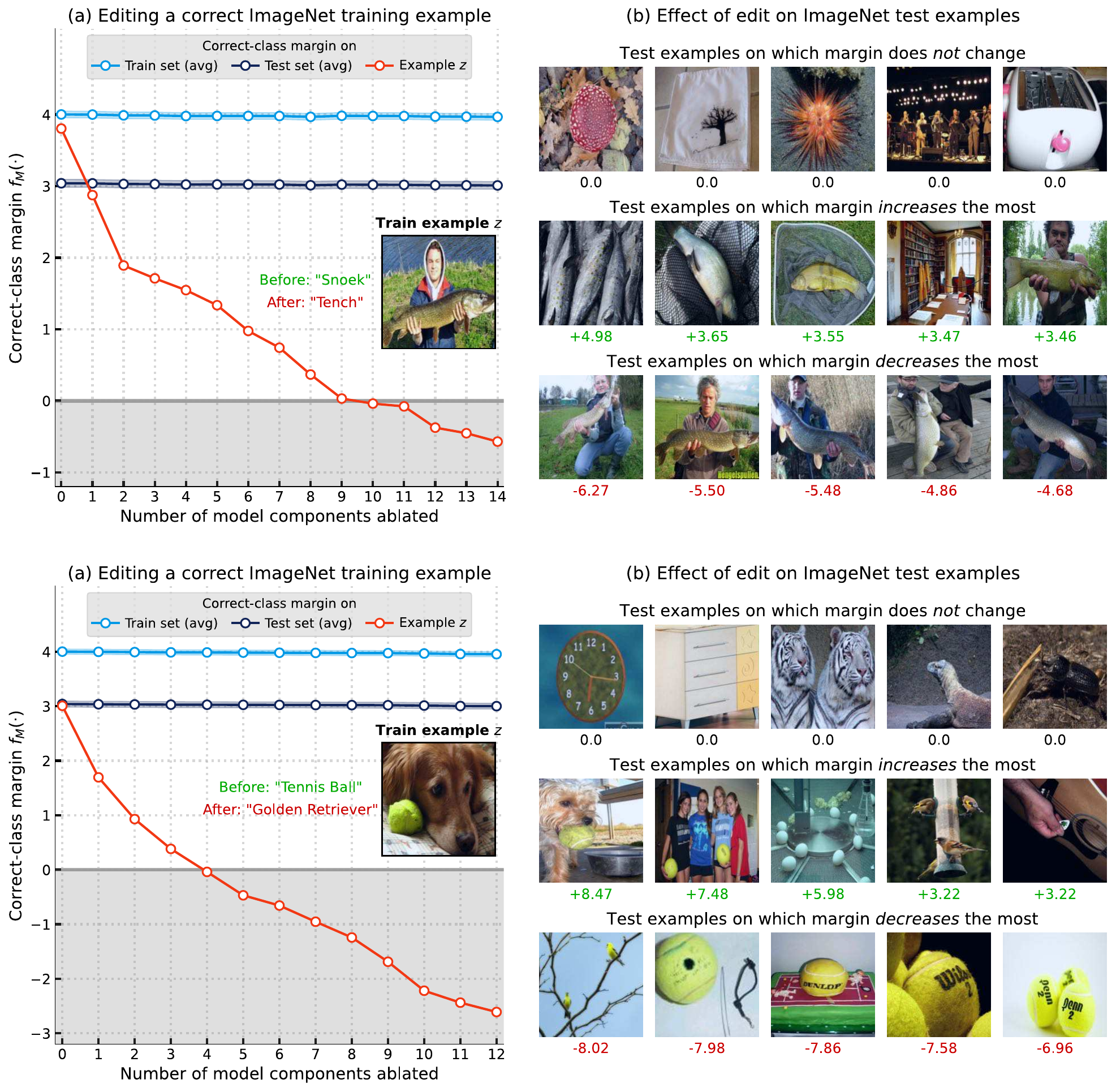}
    \caption{
        \textbf{Adversarially inducing misclassifications on ImageNet examples via \editsc.}
        Each row corresponds to an ImageNet test example that is correctly classified by a ResNet-50 model.
        In the left subplot of each row, we show that applying \editsc (by ablating the top-$k$ components \eqref{eq:c-edit}) decreases the correct-class margin \eqref{eq:margin} of the correctly classified example without impacting the average margin over the train or test set.
        On the right, we shw that the edit does not impact visually dissimilar examples, but does increase or decrease the correct-class margin of examples containing visually similar objects, e.g., tennis balls in the second row.
    }
    \label{fig:editing_in_topk}
\end{figure}

\paragraph{Additional experiments.}
We first show that \editsc is not sensitive to the choice of misclassified examples, model, or dataset.
In~\Cref{fig:editing_in_bottomk}, we reproduce the experiment in~\Cref{subsec:examples} on additional ImageNet examples misclassified by a ResNet-50 model.
In~\Cref{fig:editing_cifar10_bottomk}, we use \editsc to similarly fix misclassifications of a ResNet-18 model on the CIFAR-10 dataset.
In~\Cref{fig:editing_in_topk}, we show that \editsc can also be used to adversarially induce misclassifications on ImageNet examples by ablating the top-$k$ components corresponding to the ``target'' example.
Similar to our findings in~\Cref{subsec:examples},
we observe that ablating a few components via \editsc is sufficient to change the individual example-level prediction without changing overall model performance.

\begin{figure}[!t]
    \centering
    \includegraphics[width=0.75\textwidth]{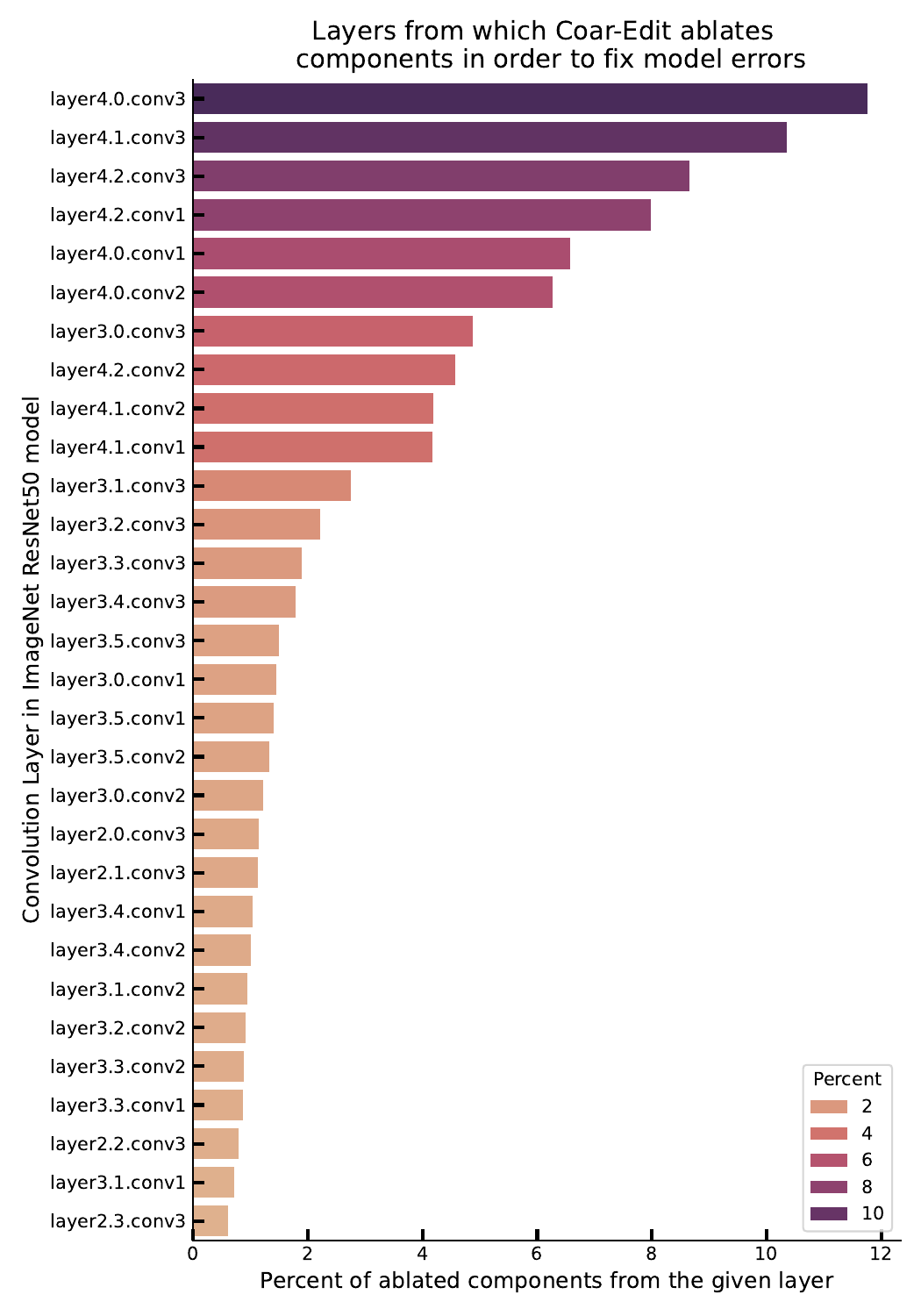}
    \caption{
        \textbf{Which components does \editsc target to fix model errors?}
        We analyze the specific convolution layers from which \editsc ablates components (convolution filters) to correct ImageNet examples misclassified by a ResNet-50 model.
        On the $y$-axis, we plot the $30$ most common convolution layers corresponding to the ablated components.
        On the $x$-axis, we plot the percentage of ablated components that belong to each convolution layer.
        We find that \editsc primarily targets convolution filters from the last few layers (closet to the output) of the ResNet-50 model in order to make fine-grained edits that do not impact overall model performance.
        For example, more than $25\%$ of the ablated components belong to \texttt{layer4.\{0,1,2\}.conv3}---the last convolution layer in the first three residual blocks of the last layer group of the ResNet-50 model.
    }
    \label{fig:whats_edited}
\end{figure}

\paragraph{Additional analysis.}
Which components does \editsc ablate to correct misclassifications?
To answer this question, we first aggregate all components ablated by \editsc in order to (individually) correct ImageNet examples misclassified by a ResNet-50 model.
Then, we plot the most common convolution layers corresponding to these ablated components in \Cref{fig:whats_edited}.
We find that \editsc primarily targets convolution filters from the last few layers (closet to the output) of the ResNet-50 model in order to make fine-grained edits that do not impact overall model performance.
For example, more than $25\%$ of the ablated components belong to \texttt{layer4.\{0,1,2\}.conv3}---the last convolution layer in the first three residual blocks of the last layer group of the ResNet-50 model.

\subsection{Forgetting a class}
\label{app:class_level}

\paragraph*{Experiment details.}
In~\Cref{subsec:class_forget}, we use \editsc to selectively forget a class of a ResNet-50 model on ImageNet.
In this experiment, we set the ``target'' examples to be set of $10$ examples from the class to be forgotten and the ``reference'' examples to be a set of $50$ randomly selected ImageNet examples.
Using these examples, we use \editsc to ablate components \eqref{eq:c-edit} that decrease the average correct-class margin \eqref{eq:margin} of the target examples without impacting the average margin over the reference examples.

\begin{figure}[!b]
    \centering
    \includegraphics[width=\textwidth]{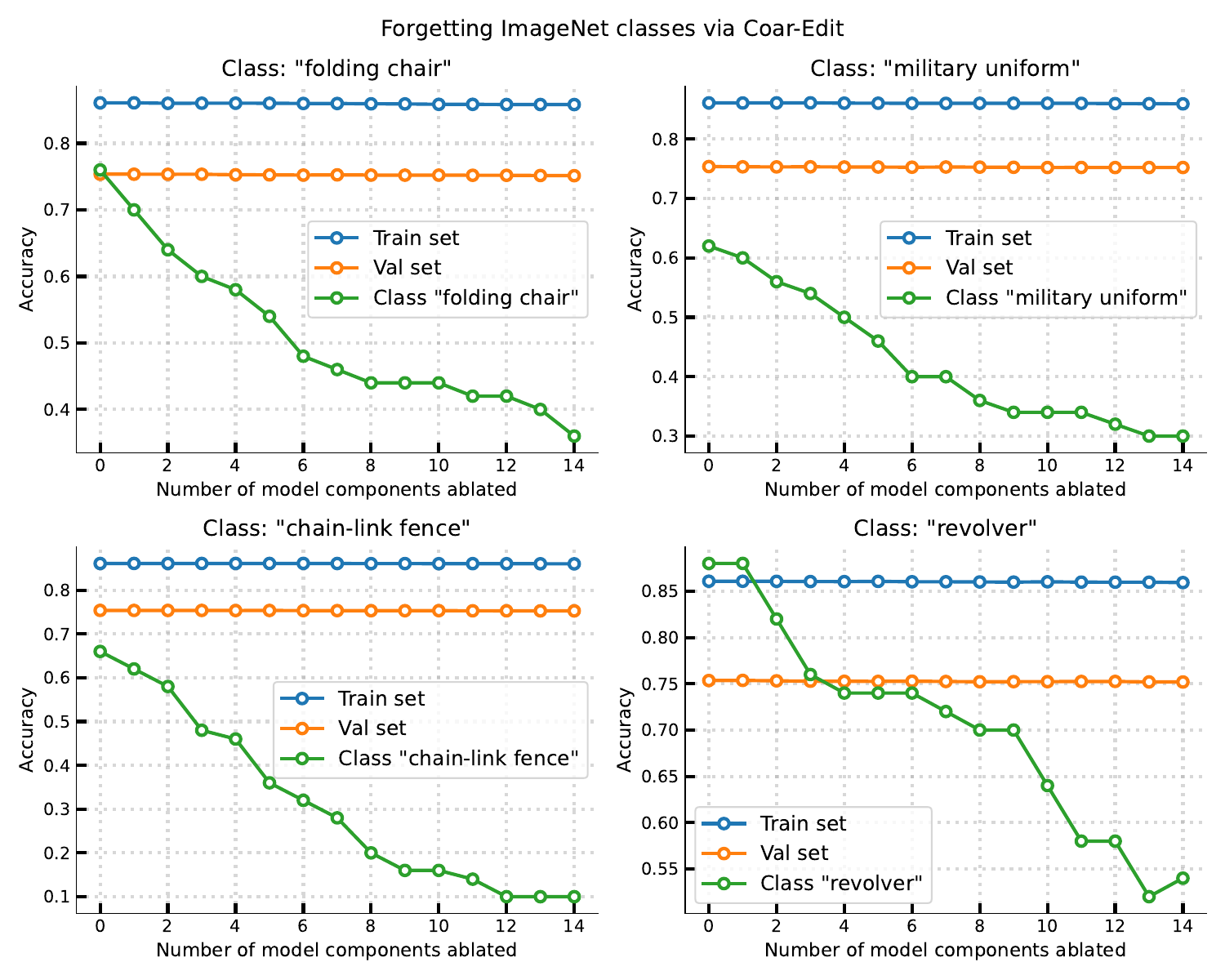}
    \caption{
        \textbf{Forgetting ImageNet classes via \editsc.}
        We reproduce the \editsc experiment from~\Cref{subsec:class_forget} on  additional ImageNet classes (one per subplot).
        Specifically, in each subplot, we find that ablating $15$ of $22,720$ convolution filters (identified via \editsc) suffices to significantly degrade the accuracy of a ResNet-50 model on a specific class (in green).
        This edit is targeted in that it does not impact the average accuracy over the train set (in blue) or test set (in orange).
    }
    \label{fig:forgetting_in_classes}
\end{figure}

\paragraph*{Additional experiments}
We show that \editsc can be used to selectively forget additional ImageNet classes.
Specifically, in~\Cref{fig:forgetting_in_classes}, we reproduce the \editsc experiment in~\Cref{subsec:class_forget} on three additional ImageNet classes: ``folding chair'', ``military uniform'', and ``revolver''.
Like in~\Cref{fig:class}, we again observe that \editsc can specifically degrade the accuracy on the target class without impacting the average accuracy over the train or test set by ablating a few components (convolution filters) in the ResNet-50 model.

\subsection{Improving subpopulation robustness.}
\label{app:group_level}

\paragraph*{Experiment details.}
In~\Cref{subsec:subpops}, we use \editsc to improve subpopulation robustness of models trained on two benchmark datasets: Waterbirds and CelebA.
In both cases, we fine-tune an ResNet-50 model via standard ``empirical risk minimization'' using SGD hyperparameters taken from \citet{sagawa2020distributionally}.
The resulting fine-tuned models attain $64\%$ and $47\%$ worst-subpopulation accuracy on the Waterbirds and CelebA test sets, respectively.
To improve subpopulation robustness on Waterbirds, we set the ``target'' examples to a set of $10$ random training examples from the ``waterbirds on land'' (the worst-performing subpopulation) and the ``reference'' examples to be $10$ random examples from other subpopulations.
Analogously, for CelebA, we set the ``target'' examples to the set of $20$ random examples from the ``blond male'' worst-performing subpopulation and the ``reference'' examples to $20$ random examples from other subpopulations.
Then, we use \editsc to identify components that, when ablated, increase the average correct-class margin \eqref{eq:margin} of the target examples without impacting the average margin over the reference examples.
In both cases, the number of components to ablate is a hyperparameter that we select by tracking the worst-subpopulation accuracy on a validation set.

\subsection{Mitigating backdoor attacks.}
\label{app:backdoor}

\paragraph{Experiment details.}
We now describe the experiment setup in~\Cref{subsec:backdoor}, where we used \editsc to mitigate the effect of a backdoor attack on a ResNet-18 model trained on a backdoored CIFAR-10 dataset.
The CIFAR-10 dataset is modified by adding a small blue-squared trigger to the upper left corner of $50\%$ of examples in the ``airplane'' class.
Training a model with standard SGD hyperparameter on this dataset causes the model to spuriously associate the trigger with the ``airplane'' class, leading to a backdoor attack.
That is, while the resulting model attains $89\%$ test accuracy, applying the attack to examples in the test set causes the model to misclassify them as ``airplanes'', resulting in $37\%$ accuracy on test examples with the trigger.
To mitigate the effect of the backdoor attack, we first sample ten examples from the training set. Then, we set the ``target'' examples to these two examples \emph{with} the trigger and the ``reference'' examples to these two examples \emph{without} the trigger.
Then, we use \editsc to ablate components \eqref{eq:c-edit} that increase the correct-class margin \eqref{eq:margin} of the target examples without impacting the average margin over the reference examples.

\begin{figure}[!t]
    \centering
    \includegraphics[width=0.7\textwidth]{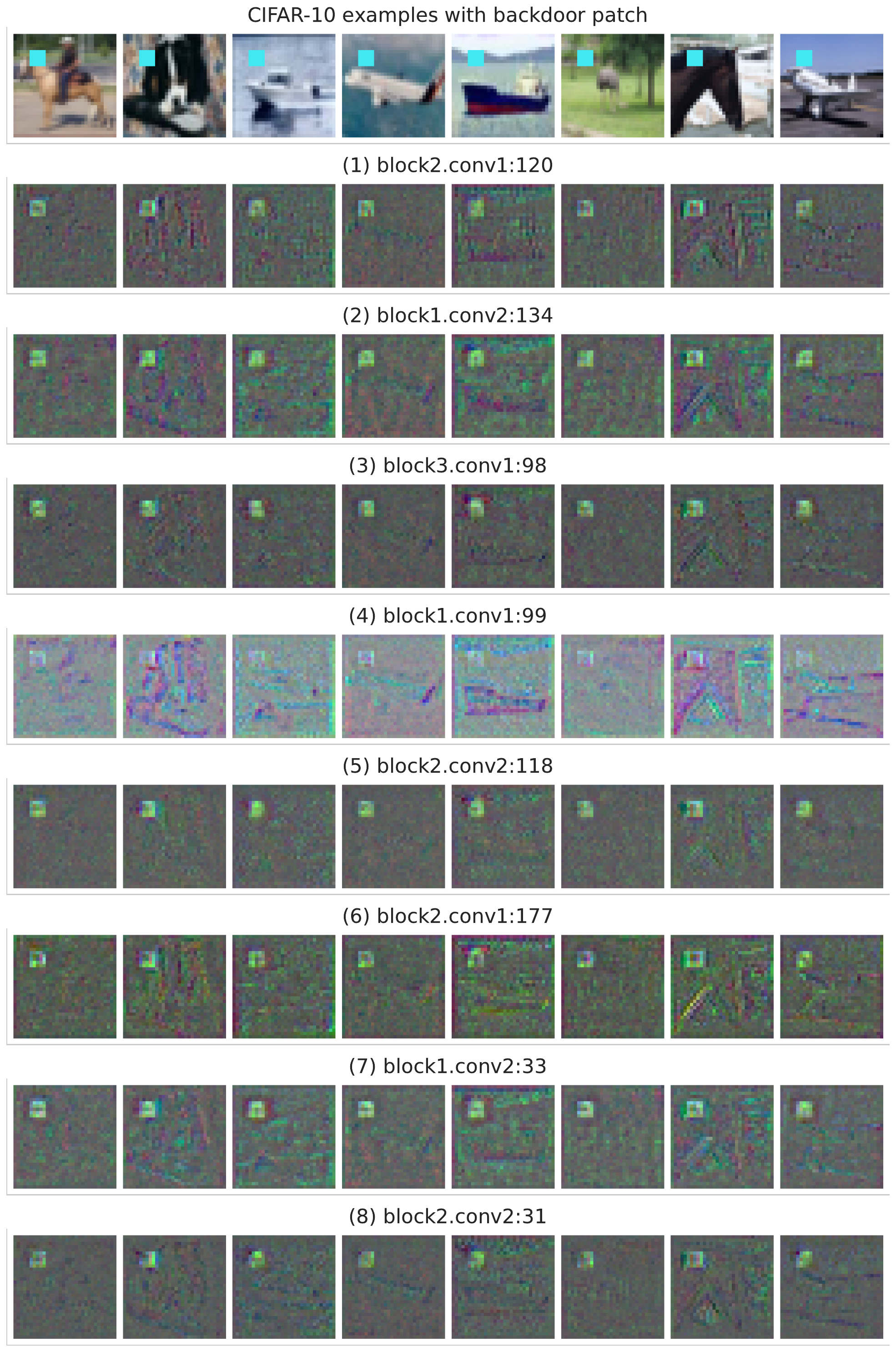}
    \caption{
        \textbf{Visualizing components ablated via \editsc to mitigate a backdoor attack.}
        Recall that in \Cref{subsec:backdoor}, we used \editsc to mitigate the effect of a backdoor attack (a blue-squared spurious trigger) on a ResNet-18 model trained on a backdoored CIFAR-10 dataset.
        Here, we visualize the components ablated via \editsc to reduce the model's reliance on this spurious feature.
        The first row shows a set of random examples from the modified CIFAR-10 test set that contain the trigger.
        Each subsequent row corresponds to an ablated component---a convolution filter of the ResNet-18 model in this case.
        In each of these rows, we use the input-times-gradient saliency map method from the Captum library \cite{kokhlikyan2020captum} to (qualitatively) highlight parts of the examples that are most ``important'' for the ablated component's output.
        These maps suggest that all ablated components are sensitive to the blue-squared trigger.
    }
    \label{fig:viz_backdoor}
\end{figure}

\paragraph{Additional analysis.}
Recall that our experiment in~\Cref{subsec:backdoor} shows that \editsc can significantly mitigate the effect of a backdoor attack on a ResNet-18 model by ablating a few backdoor-specific components.
We now qualitatively analyze the components ablated via \editsc to mitigate the effect of a backdoor attack in~\Cref{fig:viz_backdoor}.
Specifically, we visualize the ablated components (convolution filters in this case) using the input-times-gradient saliency map method from the Captum library \cite{kokhlikyan2020captum}.
As shown in~\Cref{fig:viz_backdoor}, these visualizations suggest that the ablated components are sensitive to the blue-squared trigger.

\subsection{Improving robustness to typographic attacks.}
\label{app:typographic}

\paragraph{Experiment details.}
In~\Cref{subsec:typographic} and \Cref{fig:concept_clip} in particular, we show that  \editsc can be used to improve robustness of zero-shot CLIP classifiers to typographic attacks.
In this experiment, we consider a zero-shot CLIP ViT-B/16 classifier~\cite{radford2021learning} and specify a computation graph over $82,944$ components, where each component corresponds to a weight vector in the ViT (across all layers).
We evaluate the robustness of this model in a zero-shot setting on $180$ images and four real-world typographic attacks---``taxi'', ``twitter'', ``EU'', and ``iPad''---taken from the dataset in~\cite{materzynska2022disentangling}.
We also consider synthetic typographic attacks, where we render a blob of text on a white background and place it in the center of a given image.
The zero-shot performance of the CLIP model drops from $95\%$ to $51\%$ and $54\%$ on the real and synthetic typographic attacks, respectively.
To improve robustness, we set the ``target'' examples to be the $25$ examples \emph{with} a randomly picked synthetic attack and the ``reference'' examples to the same set of examples \emph{without} any attack.
Then, we use \editsc to ablate components \eqref{eq:c-edit} that increase the average correct-class margin \eqref{eq:margin} of the target examples without impacting the average margin over the reference examples.
We use a validation set comprising examples with and without the synthetic attack to select the number of components to ablate from the model.

\section{Analyzing design choices in \methodsc}
\label{app:design_choices}
In this section, we analyze three design choices in \methodsc: (a) the train-time ablation fraction $\alpha$ used to sample a subset of components $C' \subset C$ of size $\alpha |C|$, (b) the ablation method (\Cref{rem:ablation}) used to intervene on the sampled components $C'$, and (c) the specific model output function used to compute component counterfactuals $f_M(\cdot, C')$ \eqref{eq:ablation}, i.e., model output $f_M(\cdot)$  after ablating the component subset $C'$.

\subsection{Effect of ablation fraction}
\label{app:alpha}
The first step of \methodsc---constructing a component dataset (\Cref{eq:dataset})---requires choosing a ablation fraction $\alpha \in (0, 1)$. This hyperparameter determines the size of the random $\alpha$-fraction subsets $C' \subset C$ used to compute component counterfactuals.
A priori, however, it is not clear which ablation fraction $\alpha$ is best suited for learning accurate component attributions.
For example, ablating too large a component subset (large $\alpha$) can induce a significant drop in model performance to a point where the ablated model is no longer representative of the original model.

\paragraph*{Effect of train-time ablation fraction $\alpha_{\text{train}}$}
We use two metrics to quantify the effect of ablation fraction $\alpha$ on model outputs:
\begin{itemize}
    \item \textbf{Change in model performance.} We measure the effect of ablating random $\alpha$-fraction subsets $C' \subset C$ of components on model performance, e.g., test accuracy.
    \item \textbf{Correlation between example-level model outputs.} We measure the correlation between model outputs before and after ablation, e.g., logits or margins.
\end{itemize}
We use these (heuristic) metrics to ensure that the ablations are not too severe to nullify model performance and that the outputs of the ablated models are still predictive of the outputs of the original model.

\begin{figure}[!b]
    \centering
    \includegraphics[width=\textwidth]{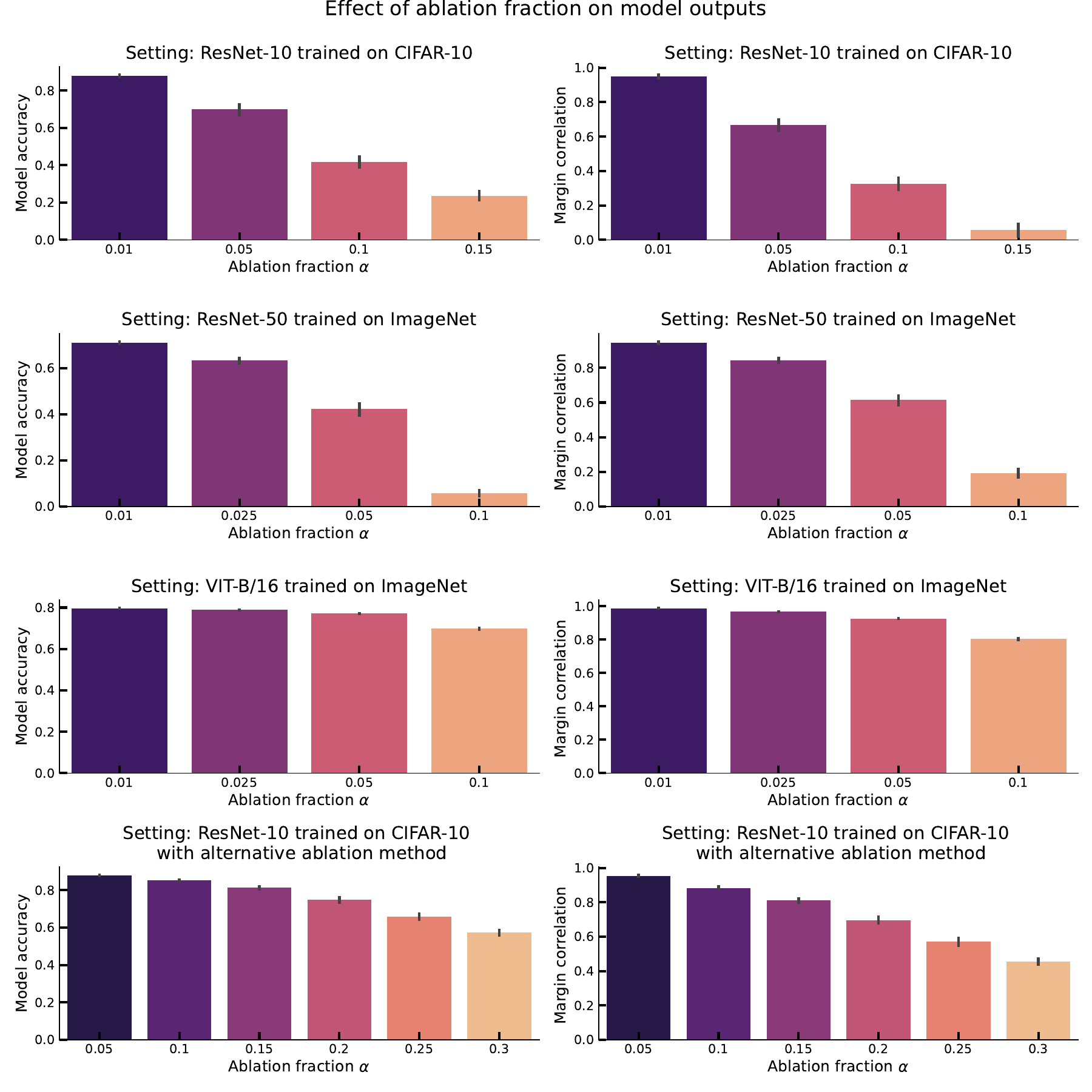}
    \caption{
        \textbf{Effect of ablation fraction $\alpha$ on model outputs.}
        We evaluate the effect of ablating $\alpha$-fraction subsets $C' \subset C$ of components ($x$-axis) on model accuracy ($y$-axis in the left column) and the correlation between model outputs before and after ablation ($y$-axis in the right column).
        In all settings considered in~\Cref{sec:eval} (one per row), we find that model accuracy and margin correlation gradually decrease as the ablation fraction $\alpha$ increases.
        See \Cref{app:alpha} for more details.
        }
    \label{fig:alpha_effect}
\end{figure}

\paragraph*{Effect of train-time ablation fraction $\alpha_{\text{train}}$.}
\Cref{fig:alpha_effect} evaluates how varying the train-time ablation fraction $\alpha_{\text{train}}$ changes both metrics---model performance and correlation between model outputs---for all three settings considered in \Cref{sec:eval}: CIFAR-10 ResNet-18, ImageNet ResNet-50, and ImageNet ViT-B/16.
In all three settings, we find that model accuracy and margin correlation decrease as the ablation fraction $\alpha$ increases.
For instance, ablating $15\%$ of components ($\alpha=0.15$) results in a significant accuracy drop for ResNets, but not for ViTs.
On the other hand, ablating $1\%$ of all components ($\alpha=0.01$) results in a small drop in accuracy and correlation, e.g., for the ResNet-18 model trained on CIFAR-10 (first row of \Cref{fig:alpha_effect}).
Therefore, our experiments in~\Cref{sec:eval} use $\alpha=0.10$ for the CIFAR-10 model and $\alpha=0.05$ for both ImageNet models.
These findings also suggest that the choice of $\alpha$ depends on the model architecture and the task at hand, e.g., ViTs are more robust to zero ablations than ResNets.

\subsection{Effect of ablation method}
\label{app:ablation_and_model_output}
As discussed in \Cref{rem:ablation}, we use a simple ablation method that sets the weights/activations of a subset of components $C' \subset C$ to zero.
However, our method \methodsc is not dependent on any specific ablation method, and can be used to compute component attributions with other ablation methods as well.

\paragraph{Alternative ablation method based on scaling.}
In this section, we consider an alternative ablation method that scales down the activations of a component by a factor of $\gamma \in [0, 1]$.
Note that setting $\gamma=0$ corresponds to the zero ablation method described in \Cref{rem:ablation}; we use $\gamma=0.5$ in our experiments.

\begin{figure}[!b]
    \centering
    \includegraphics[width=\textwidth]{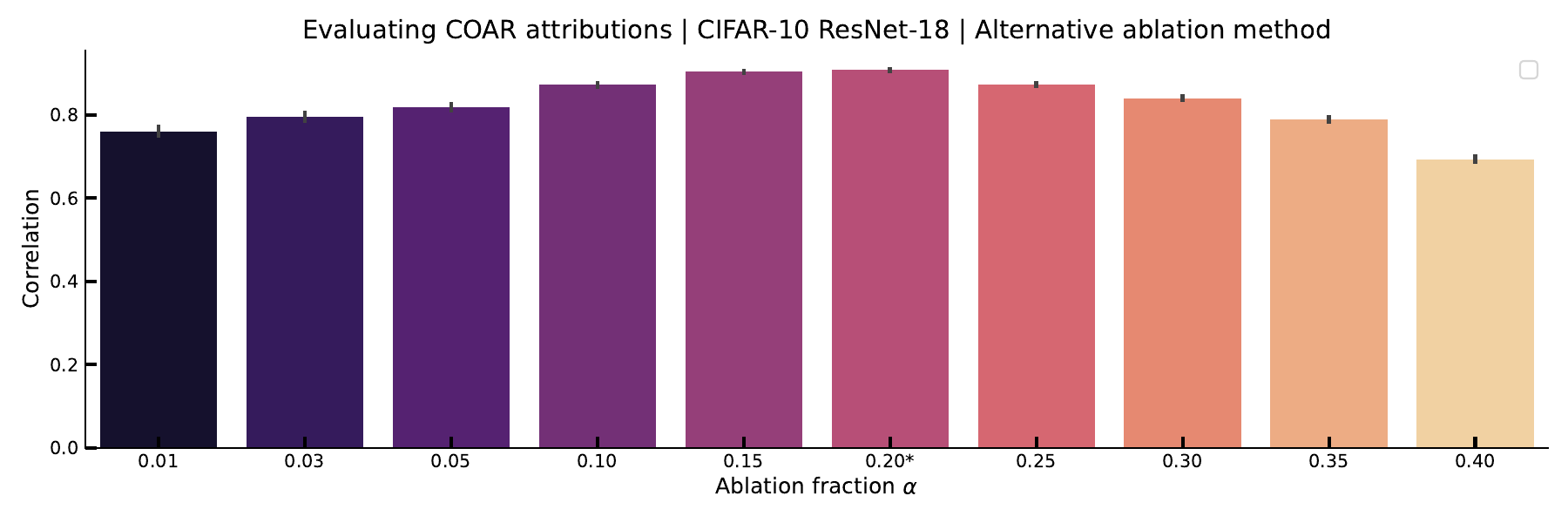}
    \caption{
        \textbf{Effect of ablation method on \methodsc attributions.}
        We estimate \methodsc attributions for a CIFAR-10 ResNet-18 model using an alternative ablation method that scales down the activations of a subset of components $C' \subset C$ by a factor of $\gamma$ ($0.5$ in this case) instead of setting them to zero.
        The resulting attribution-derived estimates \eqref{eq:attribution_derived} exhibit high correlation ($y$-axis) with ground-truth component counterfactuals.
        See \Cref{app:ablation_and_model_output} for more details.
    }
    \label{fig:alt_ablation_eval}
\end{figure}

\begin{figure}[!b]
    \centering
    \includegraphics[width=\textwidth]{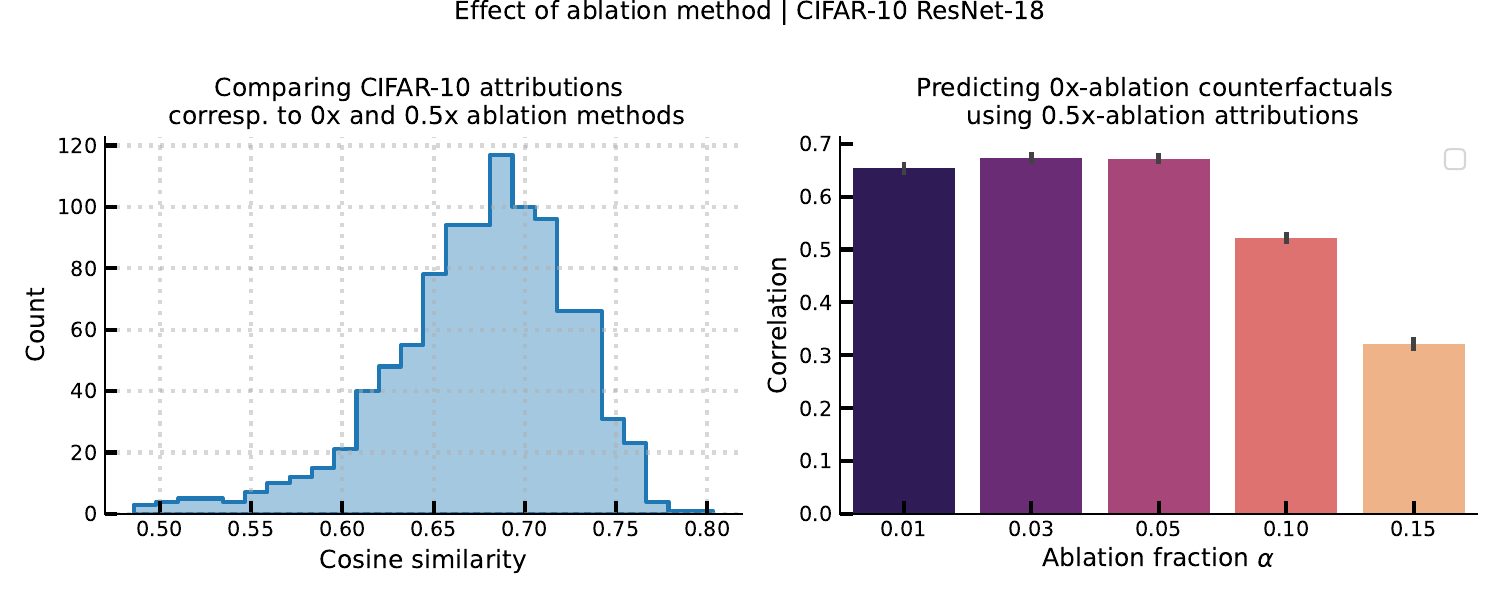}
    \caption{
        \textbf{Comparing \methodsc attributions estimated with different ablation methods.}
        We compare \methodsc attributions on a CIFAR-10 ResNet18 model computed with the zero-ablation method \Cref{rem:ablation} to attributions computed with the alternative ablation method described in \Cref{app:ablation_and_model_output}.
        The left plot shows that the paired attributions (corresponding to each example) exhibit high cosine similarity.
        The right plot shows that the counterfactual estimates \eqref{eq:attribution_derived} computed using attributions from the alternative ablation method are predictive of ground-truth component counterfactuals computed using the zero ablation method.
        See \Cref{app:ablation_and_model_output} for more details.
    }
    \label{fig:comparing_ablation}
\end{figure}

\paragraph{Experiment results.}
We find that the alternative scaling-based ablation maintains high correlation between model outputs before and after ablations, resulting in accurate component attributions.
Specifically, we make three key observations.
\begin{itemize}
    \item We first observe that on a ResNet-18 model trained on CIFAR-10, the scaling-based ablation method described above maintains high correlation between model outputs before and after ablation, even at high ablation fractions $\alpha \in \{0.30, \ldots, 0.05\}$ (fourth row of \Cref{fig:alpha_effect}). %
    \item  Then, in~\Cref{fig:alt_ablation_eval}, we apply \methodsc with the scaling-based ablation method to a CIFAR-10 ResNet-18 model. The resulting component attributions attain an average correlation of more than $0.75$ for most ablation fractions $\alpha \in \{0.40, \ldots, 0.01\}$. The correlation between \methodsc attribution estimates and ground-truth counterfactuals is high across a range of ablation fractions $\alpha$ from $0.01$ to $0.45$.
    \item In~\Cref{fig:comparing_ablation}, we compare \methodsc attributions computed with the scaling ablations to attributions computed with zero-ablations. We find that (a) these attributions exhibit high cosine similarity (\Cref{fig:comparing_ablation}a) and that (b) attributions learned with scaling-based ablations are predictive of ground-truth component counterfactuals computed using zero-ablations (\Cref{fig:comparing_ablation}b). This indicates that both ablations---scaling down the activations of a component by a factor of $\gamma=0.5$ and setting the activations of a component to zero---change model outputs in a similar manner.
\end{itemize}

\subsection{Effect of model output function}
\label{app:model_output_fn}
Recall that in~\Cref{sec:eval}, we use the correct-class margin \eqref{eq:margin} as the model output function to estimate \methodsc attributions for classification tasks.
However, our approach is not tied to a specific model output function.
Depending on the task at hand, one can use an alternative model output function to estimate \methodsc attributions.
For example, in a multi-label classification task, we can also use the logit of a fixed class of interest as the model output function to estimate \methodsc attributions.
In~\Cref{fig:tasks}, we apply \methodsc to a pre-trained ImageNet ResNet50 model fine-tuned on MIMIC-CXR~\cite{johnson2019mimic}---a dataset of labeled chest radiographs---and set the model output function to be the logit of the ``Cardiomegaly'' class.
Our results show that \methodsc attributions remain predictive with this model output function, and attain a correlation of $0.7$ and $0.6$ with the ground-truth counterfactuals on ``Cardiomegaly'' logits when $\alpha=\alpha_\text{train}=0.05$ and $\alpha=0.10$ respectively.
Additionally, in~\Cref{app:language_models}, we also apply \methodsc to the next-token prediction task in language modeling, using average correct-class margin over all tokens in a given sequence as the model output function.

\end{document}